\def\papermode{preprint}
\PassOptionsToPackage{table}{xcolor}
\documentclass{article}
\usepackage{iclr2027_conference,times}

\makeatletter
\def\tl@camera{camera}\def\tl@preprint{preprint}
\ifx\papermode\tl@camera   \iclrfinalcopy\fi
\ifx\papermode\tl@preprint \iclrfinalcopy\fi
\newcommand{\ifpreprint}[1]{\ifx\papermode\tl@preprint #1\fi}
\makeatother

\usepackage{graphicx}
\usepackage{amsmath,amssymb,amsfonts}
\usepackage{hyperref}
\hypersetup{pdftitle={Tab-Loop: Recurrent In-Context Learning for Tabular Foundation Models},colorlinks=true,linkcolor=[rgb]{0.10,0.20,0.55},citecolor=[rgb]{0.10,0.20,0.55},urlcolor=[rgb]{0.10,0.20,0.55}}
\usepackage{url}
\usepackage{booktabs}
\usepackage{multirow}
\usepackage{float}
\usepackage{subcaption}
\usepackage{xcolor}
\newcommand{\gain}{gain}

\usepackage{microtype}
\usepackage{needspace}
\usepackage{array}
\usepackage{tabularx}
\usepackage{enumitem}
\usepackage{wrapfig}
\newcommand{\method}{Tab-Loop}

\definecolor{baselinecol}{HTML}{EB6834}
\definecolor{loopcol}{HTML}{2A78D6}
\definecolor{controlcol}{HTML}{1BAF7A}
\definecolor{othercol}{HTML}{9A9A9A}
\definecolor{uniformcol}{HTML}{C2185B}
\definecolor{truncpoiscol}{HTML}{2A78D6}
\definecolor{spreadcol}{HTML}{5B3A8E}
\colorlet{baselinetint}{baselinecol!12}
\colorlet{looptint}{loopcol!12}
\colorlet{controltint}{controlcol!12}
\colorlet{uniformtint}{uniformcol!12}
\colorlet{truncpoistint}{truncpoiscol!12}
\colorlet{spreadtint}{spreadcol!12}
\newcommand{\swatch}[1]{\textcolor{#1}{\rule{0.7em}{0.7em}}}
\newcommand{\shape}[2]{\ensuremath{(#1\otimes#2)}}

\usepackage{xcolor}

\usepackage[skins,breakable]{tcolorbox}
\newtcolorbox{takehomebox}{enhanced, breakable, colback=loopcol!7, frame hidden,
  borderline west={2.5pt}{0pt}{loopcol}, sharp corners, boxrule=0pt,
  left=7pt, right=7pt, top=4pt, bottom=4pt, before skip=8pt plus 2pt, after skip=8pt plus 2pt}

\title{\method{}: Recurrent In-Context Learning for Tabular Foundation Models}

\author{
Tejas Vaidhya$^{1,2}$, Gaurav Sahu$^{1,3}$, Ishita Sanjay Hiremath$^{4}$, \& Irina Rish$^{1,2}$ \\
$^{1}$Mila -- Quebec AI Institute \\
$^{2}$Nolano AI \\
$^{3}$Unified Sciences \\
$^{4}$McGill University, Department of Bioengineering \\
\texttt{\{iamtejasvaidhya, sahu.gaurav719, ishitahiremath, irina.rish\}@gmail.com}
}

\begin{document}
\setcounter{topnumber}{2}

\maketitle
\ifpreprint{\lhead{Preprint}}

\begin{abstract}
Tabular foundation models predict unlabelled rows from labelled rows in a single forward pass via in-context learning (ICL). Like language models, these architectures have scaled by stacking transformer blocks, each with its own parameters. However, this approach couples compute with model size: increasing computation requires increasing the parameter count. In this work, we investigate whether tabular ICL requires distinct parameters at every block, or simply sufficient depth. We introduce \emph{\method{}}, a looped tabular foundation model built on the TabICLv2 architecture. Instead of stacking unique blocks, \method{} takes a small set of blocks with \emph{shared} parameters and applies them repeatedly in a \emph{loop} during its ICL stage. We evaluate this design against standard architectures under equal training and inference compute budgets. Our experiments show that depth, rather than unique parameters, drives performance. Across classification and regression tasks on held-out synthetic tables, the TabArena and TALENT benchmarks, \method{} matches the accuracy of a standard model with three times as many ICL blocks while using roughly 60\% fewer parameters. Conversely, a non-looped baseline of the same size as \method{} underperforms it, even when trained on 32\% more tables. This pattern holds consistently whether we vary the number of distinct blocks from one to twelve or the effective depth. Furthermore, running additional loops at inference improves predictive accuracy without retraining. Training the model with deeper loops extends the range over which these extra inference steps remain beneficial, eventually matching the released TabICLv2 on TabArena at 2.7 times the ICL compute. These results show that tabular foundation models can scale their compute independently of parameter count, pointing to \emph{test-time compute scaling} as a practical path toward efficient tabular models. We will open-source our model suite and all checkpoints to support future research.
\end{abstract}

\begin{figure}[!b]
  \centering
  \includegraphics[width=\linewidth]{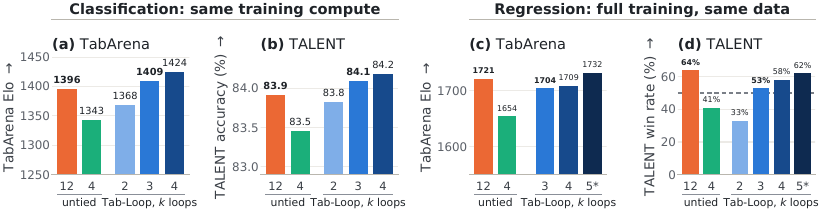}
  \caption{\textbf{\method{} matches the baseline with a third of the parameters in both tasks, and the control underperforms.} The untied baseline \shape{12}{1} applies 12 distinct blocks once; \method{} \shape{4}{k} applies 4 shared blocks for $k$ loops and matches the baseline's depth at $k=3$ (bold); the untied control \shape{4}{1} has \method{}'s parameters but a third of the depth. \textbf{(a, b)}~Classification after stage~1 at equal training compute (27.6M vs.\ 10.4M parameters): TabArena Elo (38 datasets) and TALENT accuracy (200 datasets). \textbf{(c, d)}~Regression after full training on equal data (28.5M vs.\ 11.4M): TabArena Elo (13 datasets) and TALENT win rate by RMSE.}
  \label{fig:bars}\label{fig:overview}\label{fig:overview_talent}
\end{figure}

\section{Introduction}
\label{sec:introduction}

Tabular foundation models predict the labels of a new table from its labelled rows in a single forward pass. They do so through in-context learning (ICL): the labelled rows are given to the model as context, and the model infers the relationship between features and labels without any parameter updates~\citep{hollmann2023tabpfn,hollmann2025accurate,qu2025tabicl,qu2026tabiclv2,zhang2025mitramixedsyntheticpriors,ma2025tabdpt,zhang2025limix,grinsztajn2026tabpfn3,eo2026exaone}. On small and medium-sized tables, these models now match or outperform tuned gradient-boosted trees~\citep{hollmann2025accurate,erickson2025tabarena}, which long remained the strongest approach~\citep{grinsztajn2022tree,mcelfresh2023neural}.

Like language models, tabular foundation models have gained capacity by stacking more transformer blocks, each with its own parameters. We call such a stack \emph{untied}. An untied stack ties together two quantities that need not be linked: the number of parameters a model stores, and the amount of computation it performs on each table. A model that computes for longer must therefore also be larger. This matters in practice, because a foundation model adapts at inference and its full size is paid for at every prediction~\citep{tabfm}. Model sizes are also growing quickly: TabPFN-3, for example, is about five times larger than TabPFN-2.5~\citep{grinsztajn2026tabpfn3}.

\emph{Weight sharing} removes this coupling. Instead of stacking new blocks, a model applies a small set of $n$ blocks repeatedly. We call one pass through the set a \emph{loop}, and we define the \emph{depth} $d=nk$ as the total number of block applications after $k$ loops. A looped model with 4 blocks and 3 loops therefore has the depth of a 12-block untied model, but stores only 4 blocks. In language models, looped transformers match deeper untied models on in-context tasks with substantially fewer parameters~\citep{yang2024looped,saunshi2025latent}. The idea is also natural for tabular data, where gradient boosting has long relied on applying the same kind of update many times~\citep{friedman2001greedy}.

It is not known whether a tabular foundation model can share weights across depth without losing accuracy. Prior work shows that much of the depth of these models is redundant~\citep{balef2026layerenough,koshil2026tacticl}. However, removing layers reduces depth and parameters at the same time, so it cannot reveal which of the two the model actually needs. Weight sharing reduces the number of parameters while keeping depth fixed, and therefore isolates the question: \emph{Does the ICL stage of a tabular foundation model need a distinct set of parameters for every layer, or only depth?}

We answer this question with \emph{\method{}}, a looped version of TabICLv2~\citep{qu2026tabiclv2} (Figure~\ref{fig:architecture}). We change only the ICL stage. \method{} replaces its 12 untied blocks with 4 shared blocks that are applied for $k$ loops. Before every block, it adds back the encoded table, so that later loops do not lose sight of the input. A single decoder produces predictions at every depth. During training, the number of loops is sampled at random with an average of three, which gives \method{} the same depth and the same cost per training step as TabICLv2, while it stores 62\% fewer parameters. Because the blocks are shared, the same trained model can also be run with more or fewer loops at inference.

To test whether depth or parameters matter, we compare \method{} with two untied models (Figure~\ref{fig:overview}). The \emph{baseline} has the same depth as \method{}. The \emph{control} has the same parameters as \method{} but only a third of its depth. We compare the three models at equal training compute and at equal training data. A depth--parameter scaling study of eleven smaller models, with one to twelve distinct blocks, complements this comparison. Our contributions correspond to four findings:

\begin{itemize}[leftmargin=1.2em,itemsep=3pt,topsep=3pt,parsep=0pt]
\item \textbf{Weight sharing preserves performance across depths.} In the scaling study, we keep twelve block applications and reduce the number of distinct blocks from twelve to one. Held-out loss stays identical. Applying the same blocks only once, in contrast, degrades loss, even when the compute saved is spent on additional training. Looped models also match untied models at the same depth of 24 and 48 block applications, and the same pattern appears on the 38 real datasets of TabArena-Lite, albeit with wider intervals. These experiments use a smaller training budget than the full-scale models (Section~\ref{sec:sweep}).
\item \textbf{Weight sharing preserves performance under matched compute and data.} At equal training compute, \method{} matches the baseline in both classification and regression with 62\% fewer parameters (10.4M against 27.6M). The control, although trained on 32\% more tables, underperforms in classification. When all models are trained on the same data, the equivalence persists in classification, but in regression the baseline retains an advantage on TALENT (Section~\ref{sec:matched}).
\item \textbf{Training determines the useful inference-time depth range.} When a single trained model is run with more loops, its predictions improve without retraining, up to the depths seen during training, and then level off (Section~\ref{sec:depth}).
\item \textbf{\method{} is parameter-efficient relative to existing models, at additional inference cost.} Trained on 2 to 8 loops, the 10.4M-parameter \method{} reaches 1613 Elo on TabArena, comparable to the released 27.6M-parameter TabICLv2 (1604). It remains 0.2 accuracy points below TabICLv2 on TALENT, and these results require 2.7 times the ICL compute of the baseline (Section~\ref{sec:frontier}).
\end{itemize}

\section{Why Recurrent Depth for Tabular In-Context Learning?}
\label{sec:motivation_main}

Recurrent depth lets a tabular foundation model apply its in-context learner as many times as a table requires, while storing the weights of that learner only once. Tabular foundation models have instead followed the recipe of language modelling, which is to scale: TabPFN-3 is roughly five times larger than TabPFN-2.5 ~\citep{grinsztajn2026tabpfn3}, and published models now span more than two orders of magnitude (\cite{zhang2025limix, tabfm}). Because these models adapt at inference, this growth is paid at every prediction. Compared with adding untied layers, recurrent depth has several advantages.

\begin{itemize}
[leftmargin=1.2em,itemsep=2pt,topsep=3pt,parsep=0pt]
\item \textbf{Tabular in-context learning operates iteratively over input context rather than relying on parametric memory.} Similar to how gradient boosting refines predictions by sequentially fitting residuals~\citep{friedman2001greedy}, transformer layers implement successive steps of an iterative learning algorithm~\citep{vonoswald2023,transformersbayesianinference}. Because knowledge in tabular models is drawn from labeled context rows rather than stored in parameter weights~\citep{morris2026memorize}, parameter depth becomes highly redundant. Consequently, looped architectures, where the weights are tied, can match the in-context learning capacity of much deeper untied models with substantially fewer parameters~\citep{yang2024looped,saunshi2025latent}.

\item \textbf{Existing tabular foundation models are redundant in depth.} Their layers refine representations progressively, and much of their depth can be removed~\citep{balef2026layerenough,koshil2026tacticl}. However, removing layers reduces parameters and depth together, so redundancy alone cannot say which of the two the model needs. Sharing weights reduces parameters while preserving depth, and is therefore the intervention that separates them.

\item \textbf{Recurrent depth decouples memory from computation.} A tied model stores its blocks once, so \method{} applies 12 blocks with 62\% fewer parameters than the untied baseline. Tying alone does not reduce computation, since a shared block costs as much as an untied one, but it makes computation the only cost that grows with depth.

\item \textbf{Depth becomes a choice at inference.} An untied stack cannot be shortened after training, because its decoder only virtually reads the last layer. A recurrent model with one decoder trained across depths can be read  any number of loops, and recurrent networks are known to solve harder problems by iterating longer~\citep{schwarzschild2021can,geiping2025latent}. Computation can then follow the difficulty of the dataset rather than the size of the model.

\end{itemize}

\textbf{Recurrence also carries risks, and each one shapes our design.} A shared block must perform every stage of a computation that an untied stack can spread across specialized layers~\citep{blayney2026mechanistic,bae2025relaxed}. For this reason, \method{} loops four blocks rather than one. In addition, the state of a looped model can diverge as depth grows and lose information about its input~\citep{geiping2025latent,movahedi2026fixed,yang2024looped}. We therefore add the encoded input back before every block, and we restrict recurrence to the ICL stage.

\section{\method{}}
\label{sec:method}

\method{} is TabICLv2~\citep{qu2026tabiclv2} with a single change: its ICL stage applies 4 shared blocks for $k$ loops, instead of applying 12 untied blocks once. Figure~\ref{fig:architecture} shows the three models we compare, and Table~\ref{tab:loop_vs_untied} writes out their ICL stages.

\begin{figure}[h]
\centering
\includegraphics[width=0.92\linewidth]{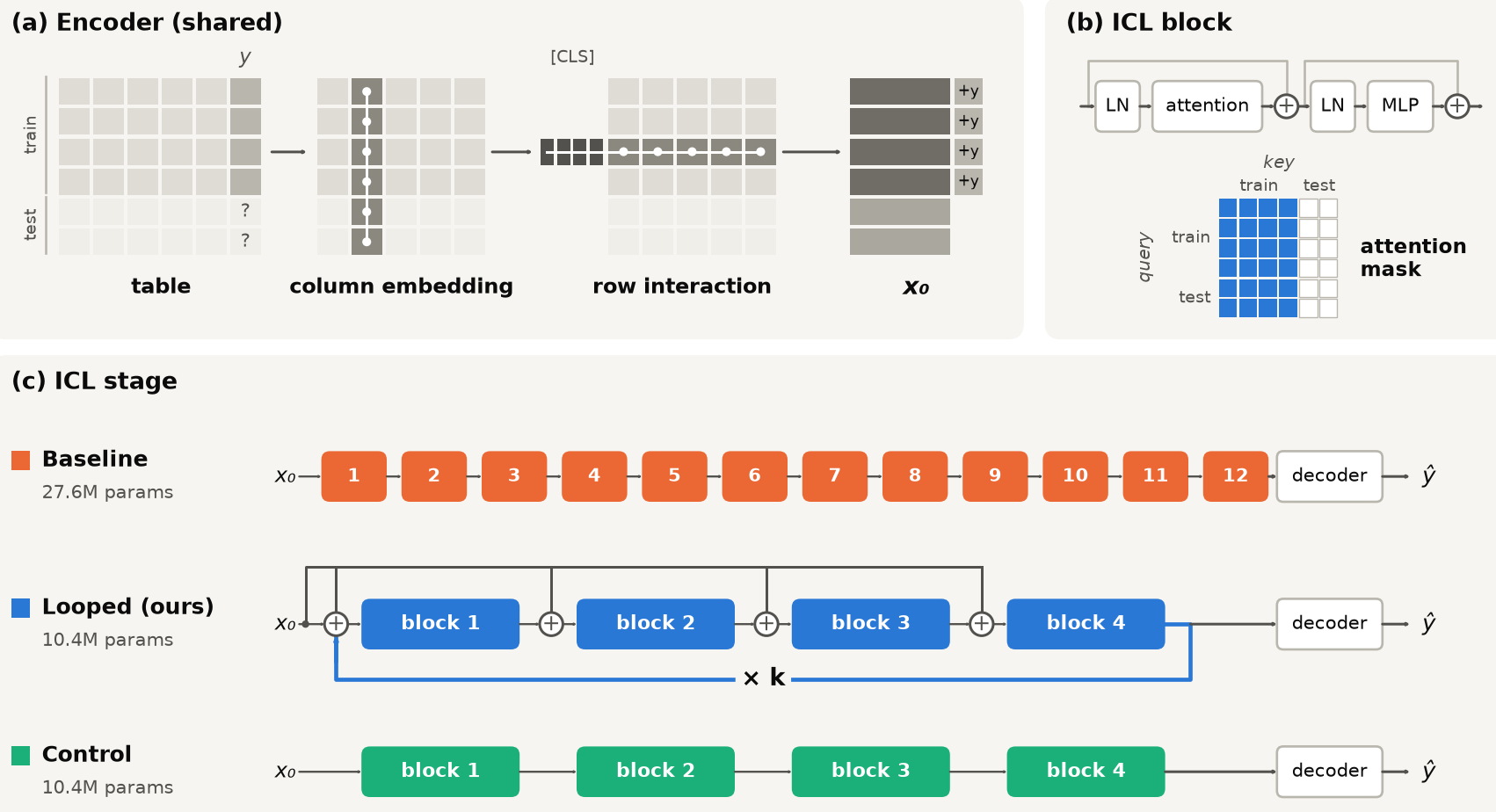}
\caption{\textbf{Models compared.} \textbf{(a)}~The column and row encoders are shared by all models and run once. \textbf{(b)}~Every ICL block is the same pre-norm transformer block; every row attends only to context rows. \textbf{(c)}~The three ICL stages (Table~\ref{tab:loop_vs_untied}); \method{} re-adds the stage input $X_0$ before every block application after the first.}
\label{fig:architecture}

\vspace{8pt}
{\small
\captionof{table}{\textbf{Untied and looped ICL stages, side by side.} All three read the encoded table $X_0$ and feed the same decoder; only the construction of depth differs. $B_\ell$: untied blocks; $C_j$: shared blocks; $S_i$: state after $i$ block applications.}
\label{tab:loop_vs_untied}
\setlength{\tabcolsep}{4pt}
\begin{tabular}{@{}lclcc@{}}
\toprule
Model & Shape & ICL stage & Applications & Params \\
\midrule
\swatch{baselinecol}\ Untied baseline & \shape{12}{1} & $H=(B_{12}\circ\cdots\circ B_{1})(X_0)$ & 12 & 27.6M \\
\swatch{controlcol}\ Untied control & \shape{4}{1} & $H=(B_{4}\circ\cdots\circ B_{1})(X_0)$ & 4 & 10.4M \\
\swatch{loopcol}\ \method{} & \shape{4}{k} & $S_{i+1}=C_{1+(i \bmod 4)}\bigl(S_i+\mathbf{1}_{\{i>0\}}X_0\bigr)$, $i<4k$ & $4k$ & 10.4M \\
\bottomrule
\end{tabular}}
\end{figure}

\subsection{Looped Architecture}
\label{sec:looped_architecture}\label{sec:principles}

\textbf{Setup.} We consider tabular classification with features $x \in \mathbb{R}^{m}$ and class labels $y \in \mathcal{Y}$. The model receives $n_{\mathrm{tr}}$ labelled context rows and predicts class probabilities for $n_{\mathrm{q}}$ query rows. Together, they form a table with $T = n_{\mathrm{tr}} + n_{\mathrm{q}}$ rows. As in TabICLv2, a column encoder first builds a representation of each cell within its column. A row encoder then combines the cells of each row into a single vector. Finally, learned label embeddings are added to the context rows, which gives the input of the ICL stage, $X_0 \in \mathbb{R}^{T \times d_{\mathrm{model}}}$. The encoders run once per forward pass. In the ICL stage, every row attends only to the context rows. Appendix Table~\ref{tab:notation} summarizes our notation.

\textbf{From a stack to a loop.} The baseline applies 12 untied blocks once, $H_{\mathrm{base}} = (B_{12} \circ \cdots \circ B_1)(X_0)$. \method{} instead uses $n$ shared blocks $C_1,\ldots,C_n$ in a fixed order. One loop applies each of them once, so $k$ loops give a depth of $d=nk$ block applications. Following \citet{saunshi2025latent}, we write \shape{n}{k} for a model with $n$ distinct blocks applied $k$ times. In this notation, the baseline is \shape{12}{1}, the control is \shape{4}{1}, and \method{} at its default depth is \shape{4}{3}. The baseline and \shape{4}{3} both have depth $d=12$. Given the input $X_0$ and $k$ loops, the state of \method{} evolves as
\begin{equation}
S_0 = X_0, \qquad
S_{i+1} = C_{1+(i \bmod n)}\left( S_i + \mathbf{1}_{\{i>0\}} X_0 \right),
\qquad i=0,\ldots,nk-1,
\label{eq:looped_icl}
\end{equation}
where $i$ counts block applications. Each block has the same attention, feed-forward, normalization and residual structure as a baseline block (Appendix~\ref{app:architecture}).

\textbf{Input re-injection.} As block applications accumulate, the state of a loop can drift away from its input~\citep{yang2024looped}. To prevent this, we add $X_0$ back before every block application after the first, including the applications within a loop. This is the indicator term in Equation~\ref{eq:looped_icl}. Every update therefore sees the encoded features and the context labels, much as an iterative solver re-reads its data at every step. The re-injection adds no parameters. Appendix~\ref{app:injection} compares it with two alternatives: no injection, and a learned linear carry, which diverges beyond the trained depths.

\textbf{Readout and cost.} A single decoder $g_\theta$ serves every depth. After $k$ loops, it reads the query rows of the final state $S_{nk}$, and for query row $t$ it predicts $p_{\theta,t}^{(k)} = \operatorname{softmax}\left( g_\theta\left([S_{nk}]_t\right) \right)$. Running $k$ loops costs $d=nk$ block applications, but only $n$ blocks are stored. With $n=4$, the total number of parameters falls from 27.6M to 10.4M. At $k=3$, \method{} and the baseline both apply 12 blocks and cost the same per table. Weight sharing therefore reduces memory, not computation; computation falls only when fewer loops are run.

\subsection{Training and Inference}
\label{sec:training}

\textbf{Three-stage pretraining.} We follow the pretraining curriculum of TabICLv2~\citep{qu2026tabiclv2}, which has three stages with increasingly large tables. Stage~1 runs for 500K steps on tables of 1{,}024 rows, stage~2 for 40K steps on 400 to 10{,}240 rows, and stage~3 for 10K steps on 400 to 60{,}000 rows with a lower learning rate (Appendix~\ref{app:pretraining}). Every model completes all three stages. For a table $\mathcal{D}$ with query rows $\mathcal{Q}_{\mathcal{D}}$, the training loss after $k$ loops is the cross-entropy on the query rows, $\ell_{\mathrm{CE}}(\theta;\mathcal{D},k)=-|\mathcal{Q}_{\mathcal{D}}|^{-1}\sum_{t\in\mathcal{Q}_{\mathcal{D}}}\log p_{\theta,t}^{(k)}[y_t]$.

\textbf{Stage 1: loop counts that average to the baseline's depth.} A model that will be run at several depths must also be trained at several depths. For a fair comparison, however, its training cost must not exceed that of the baseline. In stage~1, we therefore sample the number of loops $k$ uniformly from $\{2,3,4\}$ for each micro-batch, backpropagate through all loops, and apply the loss only at the final depth. The expected depth is $n\,\mathbb{E}_{k}[k] = 12$, the same as the baseline's, so the two models cost the same per training step. The narrow range also keeps activation memory close to the baseline's. A single loop, $k=1$, is never trained.

\textbf{Stages 2 and 3: deeper loops.} Training on two to four loops limits the depths at which the model performs well (Section~\ref{sec:depth}). To extend this range, stages~2 and~3 continue the stage-1 checkpoint under three different loop schedules, again sampled per micro-batch (Appendix Table~\ref{tab:models}). Two schedules are concentrated on 2 to 8 loops with a mean of 5: one uniform and one truncated Poisson. The third is spread out around 12 loops and also adds a latent-consistency loss. Gradients flow through the last $\min\{8,k+1\}$ loops. The baseline and the control follow the same stages. Regressors use TabICLv2's 999-quantile head with the pinball loss, and are trained with $k$ sampled from $\{2,3,4\}$ in every stage.

\section{Experimental Design}
\label{sec:design}

All full-scale models share the same prior, encoders, decoder, optimizer and learning-rate schedule. They differ only in the ICL stage. Each comparison changes a single factor and holds either the training compute or the training data fixed. We measure training compute with PyTorch's FLOP counter rather than estimating it.

\textbf{Compute-matched setting (Sections~\ref{sec:sweep} and~\ref{sec:matched}).} Every model receives the same measured training compute. A model whose training steps are cheaper therefore trains for more steps and sees more tables. This setting includes two sets of experiments. The \emph{depth--parameter scaling study} trains eleven models at 3.2~EFLOP each. Five are looped models \shape{n}{12/n} with $n\in\{1,2,3,4,6\}$, all with depth $d=12$. Six are untied models \shape{n}{1} with $n\in\{1,2,3,4,6,12\}$, which we report both at \emph{equal compute} and at \emph{equal steps} (Appendix~\ref{app:sweep_models}). Two further looped models, \shape{4}{6} and \shape{4}{12}, are compared with 24 and 48 untied blocks. The \emph{full-scale comparison} trains the baseline \shape{12}{1}, \method{} \shape{4}{3} and the control \shape{4}{1} for 18.2~EFLOP of stage-1 training each (16~EFLOP for the regressors; Table~\ref{tab:main}a). With this budget, the control trains for 660K steps and sees 32\% more tables. \method{} and the baseline differ only in weight sharing, whereas \method{} and the control differ only in depth.

\textbf{Data-matched setting (Sections~\ref{sec:matched} to~\ref{sec:frontier}).} Every model continues from its 500K-step stage-1 checkpoint through stages~2 and~3, for the same 40K and 10K steps on the same tables (Table~\ref{tab:main}b; Appendix Table~\ref{tab:models}). \method{} continues under the loop schedules described in Section~\ref{sec:training} and is evaluated at 1 to 32 loops.

\textbf{Benchmarks.} All models are applied zero-shot with TabICLv2's default preprocessing (Appendix~\ref{app:eval}). We report four measures. The first is the query log loss $L$ on 1{,}024 held-out synthetic tables; the same tables are used for every checkpoint, so comparisons are paired. The second is Elo on the 38 classification datasets and 594 splits of TabArena~\citep{erickson2025tabarena}. The third is mean test accuracy on the 200 classification datasets of TALENT~\citep{liu2024talenttabularanalyticslearning}. For regression after stage~3, we report Elo on the 13 TabArena regression datasets and win rate by RMSE on 100 TALENT datasets. We run multiple seeds and treat small differences as statistically indistinguishable.

\begin{figure}[h]
  \centering
  \includegraphics[width=\linewidth]{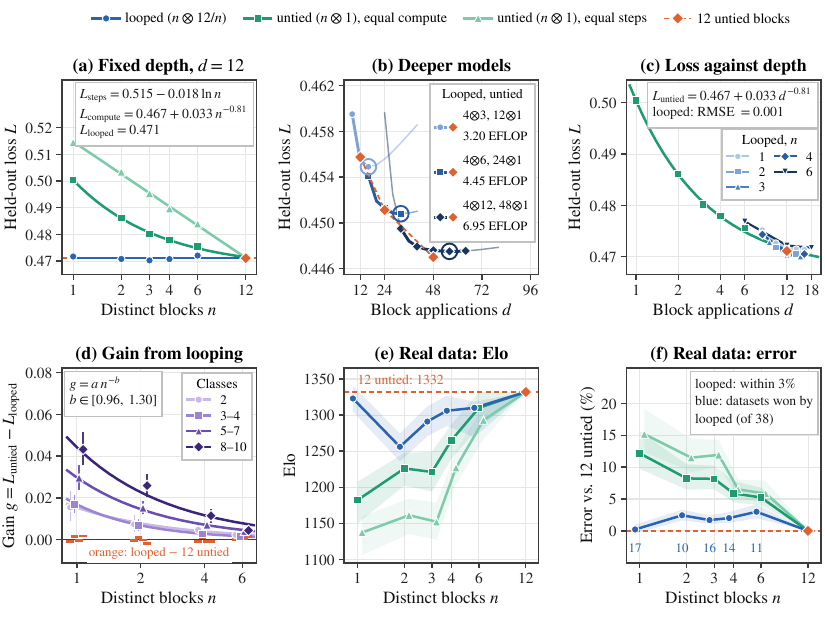}
  \caption{\textbf{Looped and untied models share one loss--depth curve, on synthetic and real tables} (depth--parameter scaling study; 3.2~EFLOP per model unless stated; 1{,}024 held-out synthetic tables; paired 95\% intervals). \textbf{(a)}~Held-out log loss against the number of distinct blocks $n$, for looped \shape{n}{12/n} and untied \shape{n}{1} models at equal compute and at equal steps; lines: fits in the legend above. \textbf{(b)}~\shape{4}{k} trained at three budgets, on $k\in\{2,3,4\}$, $\{4,\dots,8\}$ and $\{8,\dots,16\}$, at 3.20, 4.45 and 6.95~EFLOP, against untied models of the same depth; thick: trained depths; rings: best depth. This panel uses a different set of held-out tables. \textbf{(c)}~The same loss against depth $d$, with each looped model evaluated over its trained loop counts; dashed: power law fitted to the untied models only. \textbf{(d)}~The gain from looping, the loss of \shape{n}{1} minus that of \shape{n}{12/n}, by number of classes. \textbf{(e, f)}~The models of panel~(a) on TabArena-Lite (38 datasets): Elo, and geometric-mean error ratio against twelve untied blocks (numbers: datasets on which the looped model is better). Details: Appendix~\ref{app:sweep}.}
  \label{fig:sweep_analysis}\label{fig:sweep_more}
\end{figure}

\section{Results}
\label{sec:results}

\subsection{Weight Sharing Preserves Performance Across Depths}
\label{sec:sweep}\label{sec:study}\label{sec:depthnotparams}

We first ask whether reducing the number of distinct blocks hurts performance when depth is held constant. To separate the effects of parameter count and depth, we vary the number of distinct blocks while keeping the total number of block applications fixed. Figure~\ref{fig:sweep_analysis} summarizes this depth--parameter scaling study: the top row shows held-out loss on synthetic tables, while the bottom row shows how the results vary with class count and extend to real data.%made concise and removed second line-redundant

\textbf{At a fixed depth, the number of distinct blocks has no measurable effect on held-out loss.} Figure~\ref{fig:sweep_analysis}a plots held-out loss against the number of distinct blocks $n$. The blue curve shows looped models \shape{n}{12/n}, which all apply twelve blocks in total and the orange line marks twelve untied blocks. The blue curve is flat: with one to six shared blocks, held-out loss remains the same as that of twelve untied blocks. The same holds across class-count, where the loss difference between looped and untied models is negligible in every bin (orange dashes in Figure~\ref{fig:sweep_analysis}d). Even a single block applied $12$ times, \shape{1}{12}, uses one twelfth of the ICL parameters without measurable degradation.

\textbf{Reducing depth degrades held-out loss, and additional training recovers only part of it.} The two green curves in Figure~\ref{fig:sweep_analysis}a show the same $n$ blocks applied only once, \shape{n}{1}. Both rise steadily as $n$ decreases. The dotted curve compares models at equal steps: there, a single untied block is 0.043 above twelve blocks. The dashed curve gives the smaller models the compute they save as additional training steps; a single block is still  above. Additional training therefore closes only part of the gap, whereas looping the same blocks preserves performance at fixed depth. We define the \gain from looping as the loss of \shape{n}{1} minus the loss of \shape{n}{12/n}, two models with identical parameters and different depth. The gain decreases approximately as $n^{-1.16}$ (Appendix Figure~\ref{fig:sweep_price}). The bars in Figure~\ref{fig:sweep_analysis}d break it down by number of classes: they are two to three and a half times taller for tables with eight to ten classes than for binary tables, indicating a larger benefit from repeated computation on higher-class-count tables.

% \textbf{Looped and untied models follow the same loss--depth curve.} Figure~\ref{fig:sweep_analysis}c replots every model of the study against its depth $d$, the number of block applications. The green dashed curve is a power law fitted to the six untied models only, $L=0.467+0.033\,d^{-0.81}$. The blue curves, which show each looped model evaluated over its trained loop counts, closely follow this curve: across the nineteen looped readings, which were not used in the fit, the root-mean-square error is 0.001. At $d=12$, the five looped models lie between 0.470 and 0.472, whereas the loss falls by 0.029 between $d=1$ and $d=12$. The looped readings cover depths from 6 to 18, and the exponent of the fit is only loosely determined in this range (Appendix~\ref{app:sweep_fits}).

\textbf{The equivalence extends to larger models and greater depth.} Figure~\ref{fig:sweep_analysis}b shows \shape{4}{k} trained at three budgets, each evaluated over a range of loop counts, together with untied models of the same depth (triangles). The looped and untied models have nearly identical loss at matched depths. At $d=24$, \shape{4}{6} reaches 0.4513 against 0.4511 for 24 untied blocks, which have five times as many parameters and required a gentler training recipe to remain stable (Appendix~\ref{app:sweep_models}). At $d=48$, \shape{4}{12} reaches 0.4476 against 0.4470 for 48 untied blocks, with 10.4M parameters against 104.8M.

\textbf{The same pattern holds on real tables.} Figure~\ref{fig:sweep_analysis}e,f evaluates the same models on the 38 classification datasets of TabArena-Lite. Panel~(f) reports per-dataset error relative to twelve untied blocks: looped models remain within about 3\% of the reference, whereas untied models have 5--12\% higher error at equal compute and 6--15\% higher error at equal steps. Panel~(e) reports Elo across these 38 datasets. Despite this uncertainty, looped models outperform the equal-compute untied models with the same number of blocks for $n\le4$, and are indistinguishable at $n=6$. Across the fifteen models, the difference in real-data performance closely tracks the difference in held-out loss (Pearson $r=0.95$; Appendix~\ref{app:sweep_real}). These results extend the depth--parameter scaling pattern to real datasets. %just minor rewriting for consitence in text

\subsection{Weight Sharing Preserves Performance Under Matched Compute and Data}
\label{sec:matched}\label{sec:fullscale}\label{sec:regression}\label{sec:datamatched}

We next ask whether the benefit of depth is a genuine architectural effect, or whether it can be explained by differences in training compute or in the amount of data each model sees. 
%Figure~\ref{fig:overview} compares the three full-scale models, and Table~\ref{tab:main} gives the complete numbers. Its first part holds training compute fixed, and its second part holds training data fixed.
We address these possibilities through two complementary comparisons. First, we compare models trained at matched compute to establish whether weight sharing preserves the performance of a deeper baseline and whether the additional data available to the shallower control compensates for its reduced depth. Second, we hold training data fixed to determine whether the performance differences persist when all models have seen the same data.
Figure~\ref{fig:overview} summarizes the results across the three full-scale models pretraining, and Table~\ref{tab:main} reports the complete numbers, with the compute-matched and data-matched comparisons in parts (a) and (b), respectively.%giving context of what the experiment does and the results and how they flow.

\begin{table}[t]
\centering
\small
\caption{\textbf{Main results.} (a)~Stage~1 at equal compute (18.2~EFLOP per classifier, 16 per regressor). (b)~Full training, data matched: every model continues from its 500K-step stage-1 checkpoint on the same tables; $k$: loops at inference, for the schedules the best $k$ on TabArena\,/\,TALENT. Loss: held-out log loss; TabArena: Elo (regression Elo has wide intervals and is comparable only within a panel); TALENT: accuracy (\%), or win rate by RMSE in regression. $^{\dagger}$Leaderboard Elo. $^{\ddagger}$Read off Appendix Figure~\ref{fig:pareto}. Bold: best of ours.}
\label{tab:main}
\setlength{\tabcolsep}{2.6pt}
\begin{tabularx}{\linewidth}{@{}X r ccc cc@{}}
\multicolumn{7}{@{}l}{\textbf{(a) Stage 1, compute matched}} \\[1pt]
\toprule
& & \multicolumn{3}{c}{Classification} & \multicolumn{2}{c}{Regression} \\
\cmidrule(lr){3-5}\cmidrule(l){6-7}
Model & Params & Loss $\downarrow$ & TabArena$\uparrow$ & TALENT$\uparrow$ & TabArena$\uparrow$ & TALENT$\uparrow$ \\
\midrule
\swatch{baselinecol}\ Untied baseline \shape{12}{1} & 27.6/28.5M & 0.4588 & 1396 & 83.92 & 1100 & 53\% \\
\swatch{controlcol}\ Untied control \shape{4}{1} & 10.4/11.4M & 0.4681 & 1343 & 83.46 & 1082 & \textbf{56\%} \\
\rowcolor{loopcol!6}\swatch{loopcol}\ \method{} \shape{4}{3} & 10.4/11.4M & \textbf{0.4583} & \textbf{1409} & \textbf{84.10} & \textbf{1119} & \textbf{56\%} \\
% \rowcolor{loopcol!6}\swatch{loopcol}\ \method{} \shape{4}{4} & 10.4/11.4M & \textbf{0.4572} & \textbf{1424} & \textbf{84.18} & 1114 & 42\% \\
\bottomrule
\end{tabularx}

\vspace{8pt}
\begin{tabularx}{\linewidth}{@{}X r c cc cc@{}}
\multicolumn{7}{@{}l}{\textbf{(b) Full training, data matched}} \\[1pt]
\toprule
& & & \multicolumn{2}{c}{Classification} & \multicolumn{2}{c}{Regression} \\
\cmidrule(lr){4-5}\cmidrule(l){6-7}
Model & Params & $k$ & TabArena$\uparrow$ & TALENT$\uparrow$ & TabArena$\uparrow$ & TALENT$\uparrow$ \\
\midrule
\multicolumn{7}{@{}l}{\emph{Matched training; \method{} read at the baseline's depth ($k=3$, 12 block applications)}} \\
\swatch{baselinecol}\ Untied baseline \shape{12}{1} & 27.6/28.5M & -- & 1536 & 84.46 & 1721 & \textbf{64\%} \\
\swatch{controlcol}\ Untied control \shape{4}{1} & 10.4/11.4M & -- & 1470 & 83.99 & 1654 & 41\% \\
\rowcolor{loopcol!6}\swatch{loopcol}\ \method{}, $\mathcal{U}\{2,3,4\}$ & 10.4/11.4M & 3 & 1579 & 84.46 & 1704 & 53\% \\
\addlinespace[2pt]
\multicolumn{7}{@{}l}{\emph{\method{} with more loops at inference, by loop schedule in stages~2 and~3}} \\
\rowcolor{loopcol!6}\quad\swatch{loopcol}\ uniform, $\mathcal{U}\{2,3,4\}$ & 10.4/11.4M & 4 & 1586 & 84.56 & 1709 & 58\% \\
\rowcolor{loopcol!6}\quad\swatch{truncpoiscol}\ trunc.\ Poisson, $k\in\{2,\dots,8\}$ & 10.4/11.4M & 8\,/\,11;\ 5 & \textbf{1613} & 84.64 & \textbf{1732} & 62\% \\
\rowcolor{loopcol!6}\quad\swatch{uniformcol}\ uniform, $\mathcal{U}\{2,\dots,8\}$ & 10.4/11.4M & 8\,/\,9;\ \,5 & 1604 & \textbf{84.67} & 1695 & 50\% \\
% \rowcolor{loopcol!6}\quad\swatch{spreadcol}\ spread-out, $\min\{2{+}\mathrm{Pois}(10),32\}$ & 10.4M & 32 & 1565 & 84.49 & -- & -- \\
\bottomrule
\end{tabularx}
\end{table}

% \vspace{8pt}
% \begin{tabularx}{\linewidth}{@{}X r c cc cc@{}}
% \multicolumn{7}{@{}l}{\textbf{(b) Full training, data matched}} \\[1pt]
% \toprule
% & & & \multicolumn{2}{c}{Classification} & \multicolumn{2}{c}{Regression} \\
% \cmidrule(lr){4-5}\cmidrule(l){6-7}
% Model & Params & $k$ & TabArena$\uparrow$ & TALENT$\uparrow$ & TabArena$\uparrow$ & TALENT$\uparrow$ \\
% \midrule
% \multicolumn{7}{@{}l}{\emph{Matched training; \method{} read at the baseline's depth ($k=3$, 12 block applications)}} \\
% \swatch{baselinecol}\ Untied baseline \shape{12}{1} & 27.6/28.5M & -- & 1536 & 84.46 & \textbf{1728} & \textbf{64\%} \\
% \swatch{controlcol}\ Untied control \shape{4}{1} & 10.4/11.4M & -- & 1470 & 83.99 & 1661 & 41\% \\
% \rowcolor{loopcol!6}\swatch{loopcol}\ \method{}, $\mathcal{U}\{2,3,4\}$ & 10.4/11.4M & 3 & 1579 & 84.46 & 1714 & 53\% \\
% \addlinespace[2pt]
% \multicolumn{7}{@{}l}{\emph{\method{} with more loops at inference, by loop schedule in stages~2 and~3}} \\
% \rowcolor{loopcol!6}\quad\swatch{loopcol}\ uniform, $\mathcal{U}\{2,3,4\}$ & 10.4/11.4M & 4 & 1586 & 84.56 & 1716 & 58\% \\
% \rowcolor{loopcol!6}\quad\swatch{truncpoiscol}\ trunc.\ Poisson, $k\in\{2,\dots,8\}$ & 10.4M & 8\,/\,11 & \textbf{1613} & 84.64 & -- & -- \\
% \rowcolor{loopcol!6}\quad\swatch{uniformcol}\ uniform, $\mathcal{U}\{2,\dots,8\}$ & 10.4M & 8\,/\,9 & 1604 & \textbf{84.67} & -- & -- \\
% \rowcolor{loopcol!6}\quad\swatch{spreadcol}\ spread-out, $\min\{2{+}\mathrm{Pois}(10),32\}$ & 10.4M & 32 & 1565 & 84.49 & -- & -- \\
% \bottomrule
% \end{tabularx}
% \end{table}

\textbf{At equal compute, \method{} matches a baseline with three times as many ICL blocks.} We first compare \method{} against the baseline after Stage~1, when both have received the same training compute. Figure~\ref{fig:overview}a,b shows classification performance under this comparison. The bold blue bar, \method{} evaluated at three loops, has the same depth and inference cost as the orange baseline bar, and 62\% fewer parameters (10.4M against 27.6M). The two bars are level on both benchmarks: \method{} is marginally higher in TabArena Elo (1409 against 1396) and in TALENT accuracy (84.10\% against 83.92\%). Table~\ref{tab:main}a shows the same for held-out loss, which ends at 0.4583 against 0.4588 after following the baseline's throughout training (Appendix Figure~\ref{fig:heldout}). In regression, \method{} reaches 1119 Elo against 1100 and a win rate of 56\% against 53\%, within wide intervals. We interpret these differences as equivalence rather than superiority.

\textbf{Additional data does not compensate for reduced depth.} We next test whether the control's reduced depth can be compensated for by the additional training data it receives. The green bar in Figure~\ref{fig:overview}a, b is the control, which has the same parameters as \method{} but applies them only once. Because its training steps are cheaper, it sees 32\% more tables at equal compute. Despite this additional data, it is the lowest bar in both panels: 66 Elo below \method{} on TabArena and 0.6 accuracy points below on TALENT, with a higher held-out loss (0.4681; Table~\ref{tab:main}a). In regression, it is 37 Elo below on TabArena and statistically indistinguishable on TALENT. Moreover, even \method{} at two loops, the light blue bar, outperforms the control in classification.

%OLD HEADING\textbf{The equivalence persists under matched training data.} Equivalence was mentioned to be scitifcally incorrect by AI MOdels- please re-check and comment out?
\textbf{Performance is preserved under matched training data.} We next test whether the performance differences persist when all models are trained on the same data. Table~\ref{tab:main}b and Figure~\ref{fig:overview}c,d show the results after Stages~2 and~3, which all models complete on the same tables, preventing the control from converting its saved compute into additional data. In classification, \method{}, trained with a uniform loop count sampled from $\mathcal{U}(2,3,4)$ and evaluated at three loops, reaches 1579 Elo on TabArena, compared with 1536 for the baseline, while matching it on TALENT (84.46\%). The control performs worse on both benchmarks, reaching 1470 Elo on TabArena and 83.99\% on TALENT. Thus, \method{} preserves the baseline's classification performance and outperforms the control even when training data is matched. Regression performance is also comparable to baseline. In Figure~\ref{fig:overview}c, \method{} at three loops reaches 1714 Elo, close to the baseline's 1728. In Figure~\ref{fig:overview}d, the baseline wins 64\% of TALENT comparisons, compared with 53\% for \method{}. Taken together, these results show that weight sharing preserves performance relative to the deeper baseline across both classification and regression, while outperforming the shallower control in classification.

\begin{figure}[t]
  \centering
  \includegraphics[width=\linewidth]{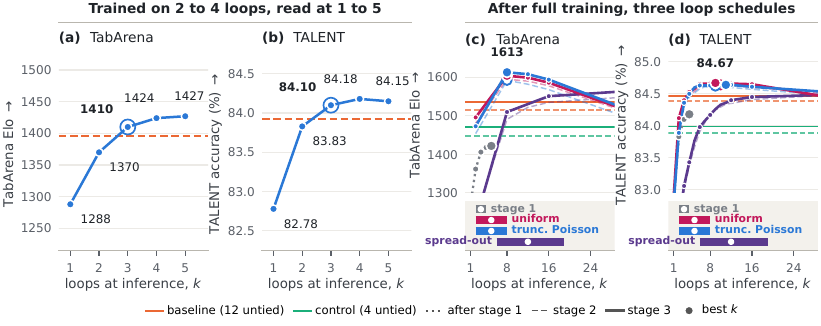}
  \caption{\textbf{Additional loops help up to the trained depths, and deeper training extends that range.} \method{} read at $k$ loops, against the untied baseline (orange) and control (green). \textbf{(a, b)}~The stage-1 checkpoint, trained on $k\in\{2,3,4\}$, on full TabArena and TALENT; ringed: $k=3$, the baseline's depth. \textbf{(c, d)}~After stages~2 and~3: the stage-1 checkpoint (grey, dotted) and the models trained further under three schedules (uniform, magenta; truncated Poisson, blue; spread-out, violet; dashed, stage~2; solid, stage~3). Large markers: peaks; bars: the loop counts each schedule trains on (white dot: mean). The spread-out model peaks at $k=32$, beyond the axis (Appendix Table~\ref{tab:schedule_peaks}).}
  \label{fig:schedule}\label{fig:test-time-depth}
\end{figure}

\subsection{Training Determines the Useful Inference-Time Depth Range}
\label{sec:depth}\label{sec:stage2}

Because the blocks are shared, the number of loops need not be fixed at training time: the same model can be evaluated with more or fewer loops at inference. We investigate this flexibility in two steps. First, we evaluate a model trained on two to four loops at different inference depths to determine how performance changes within and beyond its training range. We then vary the training loop-count schedule to test whether training on deeper loops shifts the range of useful inference depths. Figure~\ref{fig:schedule} summarizes these experiments.

\textbf{Additional loops at inference improve predictions within the trained range.} Figure~\ref{fig:schedule}a, b shows the Stage~1 checkpoint evaluated at one to five loops, with the untied baseline shown as a dashed line. Within the trained range, each additional loop improves performance: two loops come close to the baseline (1370 Elo and 83.83\%), three loops match it (1410 Elo and 84.10\%), and four loops surpass it, achieving the best results (1424 Elo and 84.18\%). At five loops, one beyond the trained range, performance plateaus rather than declining. By contrast, at  one loop, a depth never encountered during training, \method{} falls to 1288 Elo and 82.78\%, below the control, even though both models apply four blocks once. This suggests that the shared blocks distribute computation across successive loops, rather than completing it in a single pass, as in the untied model.

\textbf{Training on deeper loops shifts the range of useful inference depths} Figure~\ref{fig:schedule}c,d shows the models after Stages~2 and~3, trained under three loop-count schedules: two concentrated on 2 to 8 loops (uniform and truncated Poisson) and one spread out, with a mean of 12 loops. The bars below each panel indicate the loop counts used during training, while the large markers identify the inference depth at which each curve peaks. As training shifts toward deeper loop counts, the peak moves to greater inference depths. The Stage~1 model peaks at $k=4$ to 5, the concentrated models at $k=8$ on TabArena and $k=9$ to 11 on TALENT, and the spread-out model at $k=32$, the largest depth evaluated. Moreover, performance declines gradually beyond the peaks: the concentrated models remain at least 50 Elo above the baseline up to $k=16$, showing that the benefits of deeper training extend across a range of inference depths.
The spread-out schedule, however, does not achieve the best peak performance: its curve crosses the baseline only near $k=16$, and its peak is 48 Elo below that of the best concentrated model. Because this schedule also adds a consistency loss, the difference cannot be attributed to the loop-count distribution alone (Appendix~\ref{app:stage23}). 

\begin{figure}[t]
  \centering
  \includegraphics[width=\linewidth]{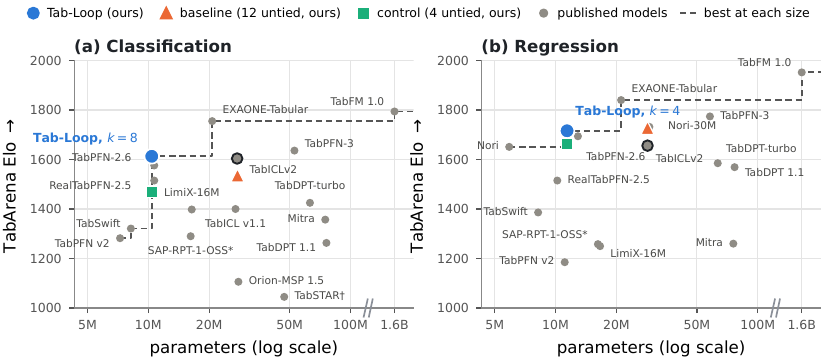}
  \caption{\textbf{\method{} lies on the frontier of Elo against parameters: no open-weight model below 20M parameters scores higher.} Full TabArena Elo against parameters in (a)~classification, with \method{} evaluated at $k=8$, and (b)~regression, at $k=4$. Grey: published models ($^{*}$gated; $^{\dagger}$with text encoder); ringed: released TabICLv2; dashed: Pareto frontier. Axes show 1000 to 2000 Elo; full range: Appendix Figure~\ref{fig:pareto}.}
  \label{fig:pareto_small}
\end{figure}

\subsection{Parameter Efficiency Relative to Existing Models}
\label{sec:frontier}

Having established how training and inference depth shape predictive performance, we now ask whether weight sharing also yields a more parameter-efficient model. Figure~\ref{fig:pareto_small} places \method{} among published open-weight models by TabArena Elo and parameter count.

\textbf{A 10.4M-parameter model matches the released TabICLv2 with substantially fewer parameters.} Figure~\ref{fig:pareto_small}a compares \method{} (blue), evaluated at eight loops, with our baseline (orange triangle) and the released TabICLv2 (ringed grey point). The concentrated models reach up to 1613 Elo, 77 points above our baseline (1536) and comparable to the released TabICLv2 (1604), while using just 38\% as many parameters. Appendix~\ref{app:replication} discusses why our baseline scores lower than the released model. This advantage also extends to TALENT: the concentrated models reach up to 84.67\%, exceeding our baseline (84.46\%) and falling just 0.2 points short of the released model (Table~\ref{tab:main}b). The dashed lines in both panels show the best published score at each model size. \method{} lies on this frontier in both classification and regression, outperforming all models with fewer than 20M parameters. EXAONE-Tabular achieves higher scores, but uses twice as many parameters.

% \textbf{The parameter savings, however, come with additional inference compute.} The horizontal axis of Figure~\ref{fig:pareto_small} measures parameter count, not computational cost. The gains above therefore reflect test-time compute scaling rather than improved parameter efficiency at equal inference cost. At $k=8$, \method{} applies $2.7\times$ as many ICL blocks as the baseline. When inference compute is matched ($k=3$), the concentrated models score up to 0.1 points below the baseline on TALENT. Moreover, the deeper schedules also increase training cost per step (Appendix~\ref{app:train_compute}), meaning that Stages~2 and~3 are not compute-matched either. Finally, each reported peak is selected separately for the benchmark on which it is reported, making these peak results somewhat optimistic (Appendix~\ref{app:eval}). Thus, weight sharing reduces the parameter count needed to achieve competitive performance, but does not necessarily reduce the computation required to obtain it.

\section{Related Work}
\label{sec:related}

\textbf{Tabular foundation models.} TabPFN~\citep{hollmann2023tabpfn} showed that a transformer pretrained on synthetic tables can classify real tables without any parameter updates~\citep{kossen2021self}. Its successors have improved the prior, the architecture and the scale~\citep{hollmann2025accurate,qu2025tabicl,qu2026tabiclv2,zhang2025mitramixedsyntheticpriors,ma2025tabdpt,priorlabs2025tabpfn25,grinsztajn2026tabpfn3,zhang2025limix,eo2026exaone,liu2026tabswift}. All of them stack untied blocks, so every increase in depth is also an increase in size. We keep the TabICLv2 recipe and change only this design choice.

\textbf{Weight sharing and recurrent depth.} Universal Transformers~\citep{dehghani2019universal} and ALBERT~\citep{lan2020albert} share parameters across layers. Looped transformers can emulate iterative algorithms~\citep{giannou2023looped}. When the input is re-injected at every step, they match standard transformers on in-context tasks with fewer parameters~\citep{yang2024looped}. They also improve reasoning and length generalization~\citep{saunshi2025latent,fan2025looped} and allow test-time compute to be scaled in language models~\citep{zhu2025scaling,geiping2025latent}, and relaxed, gated and fixed-point variants have been proposed~\citep{bae2025relaxed,hegazy2026gated,movahedi2026fixed,blayney2026mechanistic}. A related line of work on adaptive computation learns how much computation to spend on each input~\citep{graves2016adaptive,banino2021pondernet,bae2025mixture,schwarzschild2021can}. We adopt input re-injection and variable-depth evaluation from this literature and study them in a pretrained tabular model.

\textbf{Depth in tabular models.} Several works study the depth of tabular models directly. Layer reuse is effective at small scale~\citep{balef2026layerenough}, pruning can remove many blocks~\citep{koshil2026tacticl}, and early exit evaluates untied models at shallower depths~\citep{kuken2025early}. Iterative computation also appears in state-space and tiny recursive models~\citep{thielmann2024mambular,koch2025ssm,jolicoeur2025less}. Pruning and early exit, however, remove depth and parameters together, so they cannot attribute an effect to either one. In contrast, \method{} changes only the ICL stage of an established model, and we compare it with a baseline of the same depth and a control with the same parameters at equal compute.

\section{Conclusion}
\label{sec:conclusion}\label{sec:limitations}

We asked whether tabular in-context learning needs depth or parameters, and found that it needs depth. By looping a small group of weight-shared blocks inside TabICLv2, we showed that held-out loss depends on the number of block applications rather than on the number of distinct blocks: one block looped twelve times matches twelve distinct blocks, four blocks looped three times match the full model with about a third of its parameters, and extra data cannot replace the missing depth. Because one set of weights serves every depth, depth also becomes a choice at inference, so a tabular foundation model can grow its computation without growing its size.

% We asked whether the ICL stage of a tabular foundation model needs a distinct set of parameters for every layer, or only depth. To separate the two, \method{} shares four blocks across loops in the ICL stage of TabICLv2. At equal compute, it matches twelve untied blocks with 62\% fewer parameters, whereas a model with the same parameters but less depth underperforms in classification, even when trained on 32\% more tables. Across models with one to twelve distinct blocks, looped and untied models of equal depth reach the same held-out loss. Sharing blocks also makes depth a choice at inference, and training with deeper loops extends the range of depths that help. At this scale, the ICL stage therefore needs depth more than it needs distinct parameters, and the size of a tabular foundation model need not grow with the computation it performs.

% \textbf{Limitations and future work.} Our study has several limitations. We study one architecture. Training compute is matched only in stage~1. The deeper loop schedules were not run for regressors, and in regression the untied baseline still wins more often on TALENT. We did not evaluate calibration, fairness or robustness to distribution shift (Appendix~\ref{app:limitations}). A natural next step is to select the number of loops separately for each table, since the gain from looping grows with the number of classes (Appendix~\ref{app:outlook}).

\subsubsection*{Broader Impact Statement}
This work is foundational research on how tabular foundation models are scaled. Its main practical consequence is that a model with a third of the parameters can match a larger one, and that one checkpoint serves several inference budgets. This lowers the memory and storage needed to deploy tabular models and can make them easier to deploy where memory is limited. It does not lower inference compute: the gains beyond the baseline come from more loops, so they carry a higher computational and energy cost per prediction. Tabular models are widely used in high-stakes settings such as medicine, finance, hiring and public services. Higher benchmark accuracy does not imply that a model is calibrated, fair across groups or robust to distribution shift, none of which we evaluated, so deployments in such settings need the same domain-specific validation as any predictive model. Releasing our models and checkpoints supports reproducibility. We see no risks specific to looped models beyond those of tabular foundation models in general.

% \subsubsection*{AI Use Statement}
% We used Claude, an AI assistant, in preparing this paper. It drafted and edited text and restructured sections; wrote the plotting code for the figures; read numerical values from earlier versions of our figures where logged values were not at hand (these values are marked in the captions and appendix); and computed the curve fits reported in Figure~\ref{fig:sweep_analysis} and Appendix~\ref{app:sweep_fits}. All experiments were designed, run and analyzed by the authors, who checked all AI-assisted text, code, numbers and fits and take full responsibility for the content. No AI system is an author of this work.

\ificlrfinal
\subsubsection*{Acknowledgments}
We thank Mila -- Quebec AI and Nolano AI for computational resources and support.
\fi

\bibliography{references}

@inproceedings{hollmann2023tabpfn,
  title     = {{TabPFN}: A Transformer That Solves Small Tabular Classification Problems in a Second},
  author    = {Hollmann, Noah and M{\"u}ller, Samuel and Eggensperger, Katharina and Hutter, Frank},
  booktitle = {International Conference on Learning Representations},
  year      = {2023},
  url       = {https://openreview.net/forum?id=cp5PvcI6w8_}
}

@article{hollmann2025accurate,
  title   = {Accurate predictions on small data with a tabular foundation model},
  author  = {Hollmann, Noah and M{\"u}ller, Samuel and Purucker, Lennart and Krishnakumar, Arjun and K{\"o}rfer, Max and Hoo, Shi Bin and Schirrmeister, Robin Tibor and Hutter, Frank},
  journal = {Nature},
  volume  = {637},
  pages   = {319--326},
  year    = {2025},
  doi     = {10.1038/s41586-024-08328-6},
  url     = {https://www.nature.com/articles/s41586-024-08328-6}
}

@inproceedings{qu2025tabicl,
  title     = {{TabICL}: A Tabular Foundation Model for In-Context Learning on Large Data},
  author    = {Qu, Jingang and Holzm{\"u}ller, David and Varoquaux, Ga{\"e}l and {Le Morvan}, Marine},
  booktitle = {Proceedings of the 42nd International Conference on Machine Learning},
  pages     = {50817--50847},
  year      = {2025},
  volume    = {267},
  series    = {Proceedings of Machine Learning Research},
  publisher = {PMLR},
  url       = {https://proceedings.mlr.press/v267/qu25d.html}
}

@misc{qu2026tabiclv2,
  title         = {{TabICLv2}: A better, faster, scalable, and open tabular foundation model},
  author        = {Qu, Jingang and Holzm{\"u}ller, David and Varoquaux, Ga{\"e}l and {Le Morvan}, Marine},
  year          = {2026},
  eprint        = {2602.11139},
  archivePrefix = {arXiv},
  primaryClass  = {cs.LG},
  note          = {arXiv:2602.11139},
  doi           = {10.48550/arXiv.2602.11139},
  url           = {https://arxiv.org/abs/2602.11139}
}

@inproceedings{dehghani2019universal,
  title     = {Universal Transformers},
  author    = {Dehghani, Mostafa and Gouws, Stephan and Vinyals, Oriol and Uszkoreit, Jakob and Kaiser, Lukasz},
  booktitle = {International Conference on Learning Representations},
  year      = {2019},
  url       = {https://openreview.net/forum?id=HyzdRiR9Y7}
}

@inproceedings{lan2020albert,
  title     = {{ALBERT}: A Lite {BERT} for Self-supervised Learning of Language Representations},
  author    = {Lan, Zhenzhong and Chen, Mingda and Goodman, Sebastian and Gimpel, Kevin and Sharma, Piyush and Soricut, Radu},
  booktitle = {International Conference on Learning Representations},
  year      = {2020},
  url       = {https://openreview.net/forum?id=H1eA7AEtvS}
}

@inproceedings{giannou2023looped,
  title     = {Looped Transformers as Programmable Computers},
  author    = {Giannou, Angeliki and Rajput, Shashank and Sohn, Jy-Yong and Lee, Kangwook and Lee, Jason D. and Papailiopoulos, Dimitris},
  booktitle = {Proceedings of the 40th International Conference on Machine Learning},
  pages     = {11398--11442},
  year      = {2023},
  volume    = {202},
  series    = {Proceedings of Machine Learning Research},
  publisher = {PMLR},
  url       = {https://proceedings.mlr.press/v202/giannou23a.html}
}

@inproceedings{yang2024looped,
  title     = {Looped Transformers are Better at Learning Learning Algorithms},
  author    = {Yang, Liu and Lee, Kangwook and Nowak, Robert D. and Papailiopoulos, Dimitris},
  booktitle = {International Conference on Learning Representations},
  pages     = {42195--42214},
  year      = {2024},
  url       = {https://proceedings.iclr.cc/paper_files/paper/2024/hash/b8402301e7f06bdc97a31bfaa653dc32-Abstract-Conference.html}
}

@inproceedings{geiping2025latent,
  title     = {Scaling up Test-Time Compute with Latent Reasoning: A Recurrent Depth Approach},
  author    = {Geiping, Jonas and McLeish, Sean and Jain, Neel and Kirchenbauer, John and Singh, Siddharth and Bartoldson, Brian R. and Kailkhura, Bhavya and Bhatele, Abhinav and Goldstein, Tom},
  booktitle = {Advances in Neural Information Processing Systems},
  volume    = {38},
  pages     = {41340--41391},
  publisher = {Curran Associates, Inc.},
  year      = {2025},
  doi       = {10.52202/085713-1380},
  url       = {https://proceedings.neurips.cc/paper_files/paper/2025/hash/3b01972cf31e6fa0fe29e4b8b5c2a0a1-Abstract-Conference.html}
}

@inproceedings{bae2025relaxed,
  title     = {Relaxed Recursive Transformers: Effective Parameter Sharing with Layer-wise {LoRA}},
  author    = {Bae, Sangmin and Fisch, Adam and Harutyunyan, Hrayr and Ji, Ziwei and Kim, Seungyeon and Schuster, Tal},
  booktitle = {International Conference on Learning Representations},
  pages     = {34282--34327},
  year      = {2025},
  url       = {https://proceedings.iclr.cc/paper_files/paper/2025/hash/54d6a55225cebbdc16fbb0e45c5bdf2b-Abstract-Conference.html}
}

@misc{bae2025mixture,
  title         = {Mixture-of-Recursions: Learning Dynamic Recursive Depths for Adaptive Token-Level Computation},
  author        = {Bae, Sangmin and Kim, Yujin and Bayat, Reza and Kim, Sungnyun and Ha, Jiyoun and Schuster, Tal and Fisch, Adam and Harutyunyan, Hrayr and Ji, Ziwei and Courville, Aaron and Yun, Se-Young},
  year          = {2025},
  eprint        = {2507.10524},
  archivePrefix = {arXiv},
  primaryClass  = {cs.CL},
  note          = {arXiv:2507.10524},
  doi           = {10.48550/arXiv.2507.10524},
  url           = {https://arxiv.org/abs/2507.10524}
}

@inproceedings{saunshi2025latent,
  title     = {Reasoning with Latent Thoughts: On the Power of Looped Transformers},
  author    = {Saunshi, Nikunj and Dikkala, Nishanth and Li, Zhiyuan and Kumar, Sanjiv and Reddi, Sashank J.},
  booktitle = {International Conference on Learning Representations},
  pages     = {14855--14881},
  year      = {2025},
  url       = {https://proceedings.iclr.cc/paper_files/paper/2025/hash/2676109d49d1eb26d6bc584a8f556305-Abstract-Conference.html}
}

@misc{hegazy2026gated,
  title         = {Gated Recurrent Transformers: Expressive Depth through Recurrent Modulation},
  author        = {Hegazy, Amr and Alanwar, Amr and Elhoushi, Mostafa},
  year          = {2026},
  eprint        = {2608.15062},
  archivePrefix = {arXiv},
  primaryClass  = {cs.CL},
  note          = {arXiv:2608.15062},
  doi           = {10.48550/arXiv.2608.15062},
  url           = {https://arxiv.org/abs/2608.15062}
}

@inproceedings{erickson2025tabarena,
  title     = {{TabArena}: A Living Benchmark for Machine Learning on Tabular Data},
  author    = {Erickson, Nick and Purucker, Lennart and Tschalzev, Andrej and Holzm{\"u}ller, David and Desai, Prateek and Salinas, David and Hutter, Frank},
  booktitle = {Advances in Neural Information Processing Systems},
  volume    = {38},
  publisher = {Curran Associates, Inc.},
  year      = {2025},
  doi       = {10.52202/085713-0519},
  url       = {https://papers.neurips.cc/paper_files/paper/2025/hash/1697e3fb412da11dc9488249f9e7bbc9-Abstract-Datasets_and_Benchmarks_Track.html}
}

@misc{graves2016adaptive,
  title         = {Adaptive Computation Time for Recurrent Neural Networks},
  author        = {Graves, Alex},
  year          = {2016},
  eprint        = {1603.08983},
  archivePrefix = {arXiv},
  primaryClass  = {cs.NE},
  note          = {arXiv:1603.08983},
  doi           = {10.48550/arXiv.1603.08983},
  url           = {https://arxiv.org/abs/1603.08983}
}

@misc{banino2021pondernet,
  title         = {{PonderNet}: Learning to Ponder},
  author        = {Banino, Andrea and Balaguer, Jan and Blundell, Charles},
  year          = {2021},
  eprint        = {2107.05407},
  archivePrefix = {arXiv},
  primaryClass  = {cs.LG},
  note          = {arXiv:2107.05407},
  doi           = {10.48550/arXiv.2107.05407},
  url           = {https://arxiv.org/abs/2107.05407}
}

@misc{vonoswald2023,
      title={Transformers learn in-context by gradient descent}, 
      author={Johannes von Oswald and Eyvind Niklasson and Ettore Randazzo and João Sacramento and Alexander Mordvintsev and Andrey Zhmoginov and Max Vladymyrov},
      year={2023},
      eprint={2212.07677},
      archivePrefix={arXiv},
      primaryClass={cs.LG},
      url={https://arxiv.org/abs/2212.07677}, 
}

@misc{transformersbayesianinference,
      title={Transformers Can Do Bayesian Inference}, 
      author={Samuel Müller and Noah Hollmann and Sebastian Pineda Arango and Josif Grabocka and Frank Hutter},
      year={2024},
      eprint={2112.10510},
      archivePrefix={arXiv},
      primaryClass={cs.LG},
      url={https://arxiv.org/abs/2112.10510}, 
}

@misc{balef2026layerenough,
      title={Is One Layer Enough? Understanding Inference Dynamics in Tabular Foundation Models}, 
      author={Amir Rezaei Balef and Mykhailo Koshil and Katharina Eggensperger},
      year={2026},
      eprint={2605.06510},
      archivePrefix={arXiv},
      primaryClass={cs.LG},
      url={https://arxiv.org/abs/2605.06510}, 
}

@misc{liu2024talenttabularanalyticslearning,
      title={TALENT: A Tabular Analytics and Learning Toolbox}, 
      author={Si-Yang Liu and Hao-Run Cai and Qi-Le Zhou and Han-Jia Ye},
      year={2024},
      eprint={2407.04057},
      archivePrefix={arXiv},
      primaryClass={cs.LG},
      url={https://arxiv.org/abs/2407.04057}, 
}

@misc{tabfm,
      title={TabFM: A Zero-Shot Foundation Model for Tabular Data}, 
      author={Weihao Kong and Erez Louidor Ilan and Shuxin Nie and Taman Narayan and Rajat Sen and Yichen Zhou and Deqing Fu and Samet Oymak and Abhimanyu Das},
      year={2026},
      eprint={2609.37959},
      archivePrefix={arXiv},
      primaryClass={cs.LG},
      url={https://arxiv.org/abs/2609.37959}, 
}

@misc{zhang2025mitramixedsyntheticpriors,
      title={Mitra: Mixed Synthetic Priors for Enhancing Tabular Foundation Models}, 
      author={Xiyuan Zhang and Danielle C. Maddix and Junming Yin and Nick Erickson and Abdul Fatir Ansari and Boran Han and Shuai Zhang and Leman Akoglu and Christos Faloutsos and Michael W. Mahoney and Cuixiong Hu and Huzefa Rangwala and George Karypis and Bernie Wang},
      year={2025},
      eprint={2510.21204},
      archivePrefix={arXiv},
      primaryClass={cs.LG},
      url={https://arxiv.org/abs/2510.21204}, 
}

@inproceedings{ma2025tabdpt,
 author = {Ma, Junwei and Thomas, Valentin and Hosseinzadeh, Rasa and Labach, Alex and Cresswell, Jesse and Golestan, Keyvan and Yu, Guangwei and Caterini, Anthony L and Volkovs, Maks},
 booktitle = {Advances in Neural Information Processing Systems},
 doi = {10.52202/085713-5748},
 editor = {D. Belgrave and C. Zhang and H. Lin and R. Pascanu and P. Koniusz and M. Ghassemi and N. Chen},
 pages = {172692--172722},
 publisher = {Curran Associates, Inc.},
 title = {TabDPT: Scaling Tabular Foundation Models on Real Data},
 url = {https://proceedings.neurips.cc/paper_files/paper/2025/file/fc0e3f908a2116ba529ad0a1530a3675-Paper-Conference.pdf},
 volume = {38, Main Conference},
 year = {2025}
}

@article{friedman2001greedy,
  title   = {Greedy Function Approximation: A Gradient Boosting Machine},
  author  = {Friedman, Jerome H.},
  journal = {Annals of Statistics},
  pages   = {1189--1232},
  year    = {2001},
}

@inproceedings{grinsztajn2022tree,
  title     = {Why do tree-based models still outperform deep learning on typical tabular data?},
  author    = {Grinsztajn, L{\'e}o and Oyallon, Edouard and Varoquaux, Ga{\"e}l},
  booktitle = {Advances in Neural Information Processing Systems, Datasets and Benchmarks Track},
  year      = {2022},
}

@inproceedings{mcelfresh2023neural,
  title     = {When Do Neural Nets Outperform Boosted Trees on Tabular Data?},
  author    = {McElfresh, Duncan and Khandagale, Sujay and Valverde, Jonathan and Prasad C, Vishak and Ramakrishnan, Ganesh and Goldblum, Micah and White, Colin},
  booktitle = {Advances in Neural Information Processing Systems},
  pages     = {76336--76369},
  year      = {2023},
}

@inproceedings{kossen2021self,
  title     = {Self-Attention Between Datapoints: Going Beyond Individual Input-Output Pairs in Deep Learning},
  author    = {Kossen, Jannik and Band, Neil and Lyle, Clare and Gomez, Aidan N. and Rainforth, Tom and Gal, Yarin},
  booktitle = {Advances in Neural Information Processing Systems},
  volume    = {34},
  pages     = {28742--28756},
  year      = {2021},
}

@misc{priorlabs2025tabpfn25,
  title         = {{TabPFN-2.5}: Advancing the State of the Art in Tabular Foundation Models},
  author        = {{Prior Labs}},
  year          = {2025},
  eprint        = {2511.08667},
  archivePrefix = {arXiv},
  primaryClass  = {cs.LG},
}

@misc{grinsztajn2026tabpfn3,
  title         = {{TabPFN-3}: Technical Report},
  author        = {Grinsztajn, L{\'e}o and Fl{\"o}ge, Klemens and Key, Oscar and Birkel, Felix and Jund, Philipp and Roof, Brendan and Manium, Mihir and Hoo, Shi Bin and B{\"u}hler, Magnus and Garg, Anurag and others},
  year          = {2026},
  eprint        = {2605.13986},
  archivePrefix = {arXiv},
  primaryClass  = {cs.LG},
}

@misc{zhang2025limix,
  title         = {{LimiX}: Unleashing Structured-Data Modeling Capability for Generalist Intelligence},
  author        = {Zhang, Xingxuan and Ren, Gang and Yu, Han and Yuan, Hao and Wang, Hui and others},
  year          = {2025},
  eprint        = {2509.03505},
  archivePrefix = {arXiv},
  primaryClass  = {cs.LG},
}

@misc{eo2026exaone,
  title         = {{EXAONE Tabular 1.0}: Technical Report},
  author        = {Eo, Moonjung and Suh, Min-Kook and Cho, Hye-Seung and Kim, Jiwon and Kim, Seoyoon and Nam, Sangjun and Lee, Soonyoung},
  year          = {2026},
  eprint        = {2608.25774},
  archivePrefix = {arXiv},
  primaryClass  = {cs.LG},
}

@inproceedings{liu2026tabswift,
  title     = {{TabSwift}: An Efficient Tabular Foundation Model with Row-wise Attention},
  author    = {Liu, Si-Yang and Ye, Han-Jia},
  booktitle = {Proceedings of the 43rd International Conference on Machine Learning},
  series    = {Proceedings of Machine Learning Research},
  year      = {2026},
}

@misc{koshil2026tacticl,
  title         = {{TACTICL}: Task-Aware Compression of Tabular {ICL} Models},
  author        = {Koshil, Mykhailo and Feurer, Matthias and Eggensperger, Katharina},
  year          = {2026},
  eprint        = {2608.10837},
  archivePrefix = {arXiv},
  primaryClass  = {cs.LG},
}

@misc{thielmann2024mambular,
  title         = {Mambular: A Sequential Model for Tabular Deep Learning},
  author        = {Thielmann, Anton Frederik and Kumar, Manish and Weisser, Christoph and Reuter, Arik and S{\"a}fken, Benjamin and Samiee, Soheila},
  year          = {2024},
  eprint        = {2408.06291},
  archivePrefix = {arXiv},
  primaryClass  = {cs.LG},
}

@misc{koch2025ssm,
  title         = {State-Space Models for Tabular Prior-Data Fitted Networks},
  author        = {Koch, Felix and Wever, Marcel and Raisch, Fabian and Tischler, Benjamin},
  year          = {2025},
  eprint        = {2510.14573},
  archivePrefix = {arXiv},
  primaryClass  = {cs.LG},
}

@misc{jolicoeur2025less,
  title         = {Less is More: Recursive Reasoning with Tiny Networks},
  author        = {Jolicoeur-Martineau, Alexia},
  year          = {2025},
  eprint        = {2510.04871},
  archivePrefix = {arXiv},
  primaryClass  = {cs.LG},
}

@inproceedings{schwarzschild2021can,
  title     = {Can You Learn an Algorithm? Generalizing from Easy to Hard Problems with Recurrent Networks},
  author    = {Schwarzschild, Avi and Borgnia, Eitan and Gupta, Arjun and Huang, Furong and Vishkin, Uzi and Goldblum, Micah and Goldstein, Tom},
  booktitle = {Advances in Neural Information Processing Systems},
  volume    = {34},
  pages     = {6695--6706},
  year      = {2021},
}

@inproceedings{fan2025looped,
  title     = {Looped Transformers for Length Generalization},
  author    = {Fan, Ying and Du, Yilun and Ramchandran, Kannan and Lee, Kangwook},
  booktitle = {International Conference on Learning Representations},
  year      = {2025},
}

@misc{zhu2025scaling,
  title         = {Scaling Latent Reasoning via Looped Language Models},
  author        = {Zhu, Rui-Jie and Wang, Zixuan and Hua, Kai and Zhang, Tianyu and Li, Ziniu and Que, Haoran and Wei, Boyi and Wen, Zixin and Yin, Fan and Xing, He and others},
  year          = {2025},
  eprint        = {2510.25741},
  archivePrefix = {arXiv},
  primaryClass  = {cs.CL},
}

@misc{movahedi2026fixed,
  title         = {Fixed-Point Reasoners: Stable and Adaptive Deep Looped Transformers},
  author        = {Movahedi, Sajad and Milovanovi{\'c}, Vera and Feigin, Shlomo Libo and Theus, Alexander and Hofmann, Thomas and Boeva, Valentina and Rusch, T. Konstantin and Orvieto, Antonio},
  year          = {2026},
  eprint        = {2606.18206},
  archivePrefix = {arXiv},
  primaryClass  = {cs.LG},
}

@misc{blayney2026mechanistic,
  title         = {A Mechanistic Analysis of Looped Reasoning Language Models},
  author        = {Blayney, Hugh and Arroyo, {\'A}lvaro and Obando-Ceron, Johan and Castro, Pablo Samuel and Courville, Aaron and Bronstein, Michael M. and Dong, Xiaowen},
  year          = {2026},
  eprint        = {2604.11791},
  archivePrefix = {arXiv},
  primaryClass  = {cs.LG},
}

@inproceedings{kuken2025early,
  title     = {Early Stopping Tabular In-Context Learning},
  author    = {K{\"u}ken, Jaris and Purucker, Lennart and Hutter, Frank},
  booktitle = {1st International Workshop on Foundation Models for Structured Data (FMSD) at ICML},
  year      = {2025},
}

@inproceedings{morris2026memorize,
  title     = {How Much Can Language Models Memorize?},
  author    = {Morris, John Xavier and Sitawarin, Chawin and Guo, Chuan and Kokhlikyan, Narine and Suh, G. Edward and Rush, Alexander M. and Chaudhuri, Kamalika and Mahloujifar, Saeed},
  booktitle = {Proceedings of the 43rd International Conference on Machine Learning},
  year      = {2026},
}

@misc{jordan2024muon,
  author = {Jordan, Keller and Jin, Yuchen and Boza, Vlado and You, Jiacheng and Cesista, Franz and Newhouse, Laker and Bernstein, Jeremy},
  title = {Muon: An optimizer for hidden layers in neural networks},
  year = {2024},
  howpublished = {\url{https://kellerjordan.github.io/posts/muon/}}
}

@article{liu2025muon,
  author = {Liu, Jingyuan and Su, Jianlin and Yao, Xingcheng and Jiang, Zhejun and Lai, Guokun and Du, Yulun and Qin, Yidao and Xu, Weixin and Lu, Enzhe and Yan, Junjie and others},
  title = {Muon is Scalable for {LLM} Training},
  journal = {arXiv preprint arXiv:2502.16982},
  year = {2025}
}

@misc{balef2026loopicl,
      title={{LoopICL}: Looping a single transformer block to solve tabular tasks},
      author={Amir Rezaei Balef and Katharina Eggensperger},
      year={2026},
      eprint={2609.36108},
      archivePrefix={arXiv},
      primaryClass={cs.LG},
      url={https://arxiv.org/abs/2609.36108},
}
\bibliographystyle{iclr2027_conference}

\appendix
\vspace{8pt}
\raggedbottom
\setcounter{topnumber}{4}\setcounter{bottomnumber}{3}\setcounter{totalnumber}{6}
\renewcommand{\topfraction}{0.92}\renewcommand{\bottomfraction}{0.85}
\renewcommand{\textfraction}{0.06}\renewcommand{\floatpagefraction}{0.85}
\setlength{\parskip}{4pt plus 1pt minus 1pt}
\setlength{\floatsep}{10pt plus 3pt minus 2pt}
\setlength{\textfloatsep}{12pt plus 3pt minus 2pt}
\setlength{\intextsep}{10pt plus 3pt minus 2pt}
\renewcommand{\arraystretch}{1.15}
\makeatletter
\renewcommand\section{\@startsection{section}{1}{\z@}{-3.5ex plus -.6ex minus -.2ex}{2ex plus .3ex}{\Large\bfseries\raggedright}}
\renewcommand\subsection{\@startsection{subsection}{2}{\z@}{-2.8ex plus -.6ex minus -.2ex}{1.4ex plus .2ex}{\large\bfseries\raggedright}}
\renewcommand\subsubsection{\@startsection{subsubsection}{3}{\z@}{-2.4ex plus -.5ex minus -.2ex}{1ex plus .2ex}{\normalsize\bfseries\raggedright}}
\renewcommand\@seccntformat[1]{\ifnum\pdfstrcmp{#1}{section}=0 Appendix~\fi\csname the#1\endcsname\hspace{0.7em}}
\makeatother

\newpage
\section*{Appendix}
\label{app:overview}

The appendix has three parts: the models and how they were trained and evaluated (Appendices~\ref{app:architecture}--\ref{app:eval}), the additional results, with classification and regression side by side (Appendices~\ref{app:sweep}--\ref{app:depthdial}). Within each part, appendices follow the order of the main text, and each begins by naming the section it supports.

\newcommand{\apppartrow}[1]{\multicolumn{4}{@{}l}{\rule{0pt}{3.2ex}\textsc{#1}} \\[1pt] \midrule}
\newcommand{\appsub}[1]{\par\vspace{-1pt}{\footnotesize\color{black!65}#1}}
\newcommand{\apppart}[2]{\needspace{8\baselineskip}\pdfbookmark[0]{#2}{#1}\noindent{\small\scshape #2}\par\vspace{-3pt}\noindent\rule{\linewidth}{0.4pt}\par\vspace{-2.5ex}}
\newcommand{\appentry}[3]{\ref{#1} & \hyperref[#1]{\textbf{#2}} & #3 & \pageref{#1} \\}
\newcommand{\appsubrow}[2]{& {\footnotesize\hspace{1.2em}\hyperref[#1]{\color{black!70}\ref*{#1}\hspace{0.6em}#2}} & & {\footnotesize\color{black!70}\pageref*{#1}} \\}
\newcommand{\appnote}[1]{& {\footnotesize\hspace{1.2em}\color{black!70}#1} & & \\}
\newcommand{\appgap}{\addlinespace[5pt]}

\begin{center}
\small
\renewcommand{\arraystretch}{1.0}
\setlength{\tabcolsep}{5pt}
\begin{tabular}{@{}l>{\raggedright\arraybackslash}p{10.1cm}cr@{}}
\toprule
& \textbf{Appendix} & \textbf{Main text} & \textbf{Page} \\
\apppartrow{Part I\quad Models, training and evaluation}
\appentry{app:architecture}{Architecture}{\S\ref{sec:method}}
\appentry{app:pretraining}{Pretraining}{\S\ref{sec:training}}
\appentry{app:compute_full}{Compute accounting}{\S\ref{sec:design}}
\appentry{app:eval}{Evaluation protocols}{\S\ref{sec:design}}
\apppartrow{Part II\quad Results}
\appentry{app:sweep}{Depth--parameter scaling study}{\S\ref{sec:sweep}}
\appentry{app:fullscale}{Full-setup comparison, compute and data matched}{\S\ref{sec:matched}, \S\ref{sec:datamatched}}
\appentry{app:depthdial}{Test-time depth and training schedules}{\S\ref{sec:depth}}
% \apppartrow{Part III\quad Theory}
% \appentry{app:motivation_full}{A linear model of weight-shared and untied depth}{\S\ref{sec:motivation_main}, \S\ref{sec:results}}
\apppartrow{Part III\quad Discussion}
\appentry{app:limitations}{Limitations}{\S\ref{sec:conclusion}}
% \appentry{app:outlook}{Outlook: open hypotheses}{\S\ref{sec:conclusion}}
\bottomrule
\end{tabular}
\end{center}

\clearpage
\apppart{partI}{Part I\texorpdfstring{\quad}{: }Models, training and evaluation}
\section{Architecture}
\label{app:architecture}

\emph{Supports Section~\ref{sec:method}.} This appendix describes the \method{} backbone stage by stage (Appendix~\ref{app:backbone_detail}), summarizes its components and the three ICL variants (Appendix~\ref{app:arch}), and gives the parameter counts (Appendix~\ref{app:params}), the ICL block (Appendix~\ref{app:icl_block}) and the input injection used by \method{} (Appendix~\ref{app:injection}). The three models in the paper differ only in their ICL stage; everything else is shared. Table~\ref{tab:notation} collects the notation used throughout.

\subsection{The \method{} Backbone, Stage by Stage}
\label{app:backbone_detail}

\method{} inherits its backbone from TabICLv2~\citep{qu2026tabiclv2}, which refines TabICL~\citep{qu2025tabicl}. We describe it here for readers who do not know these models. The backbone turns a table into predictions in four stages (Figure~\ref{fig:backbone_stages}): it first embeds every cell using the other cells of its column, then compresses every row into one vector, then lets the unlabeled rows read the labeled ones, and finally decodes a prediction for each unlabeled row. Only the third stage, the ICL stage, differs between the models of this paper.

\begin{figure}[!t]
  \centering
  \includegraphics[width=\linewidth]{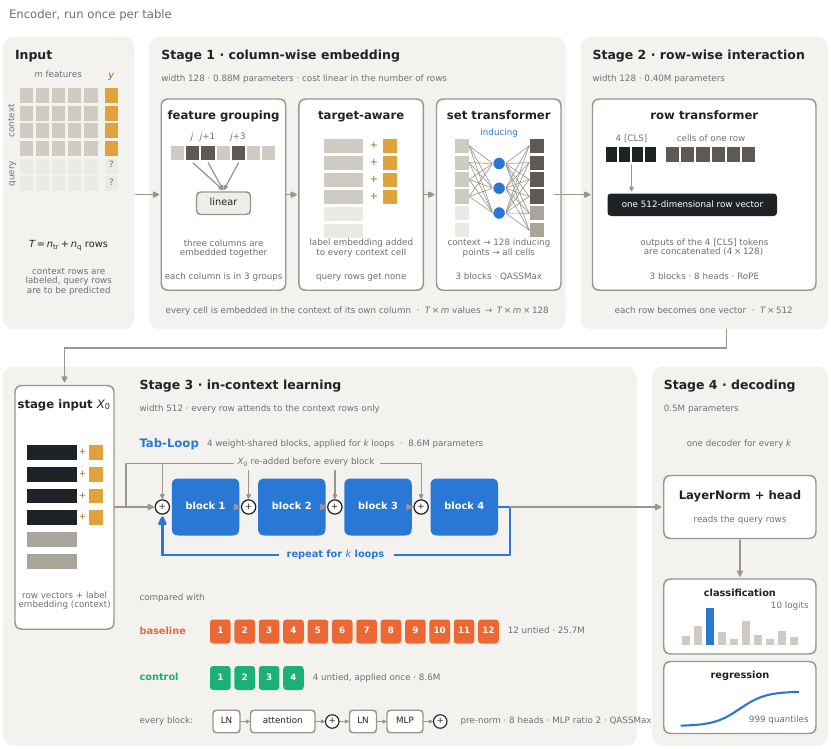}
  \caption{\textbf{The \method{} backbone, stage by stage.} Stages~1 and~2 embed the table and run once per forward pass: every cell is embedded within its column, and every row is then compressed into one vector. Stage~3 is the ICL stage, where the query rows read the context rows; \method{} applies 4 weight-shared blocks for $k$ loops and adds the stage input $X_0$ back before every block application after the first, whereas the untied baseline applies 12 distinct blocks once. Stage~4 decodes a prediction for each query row, and the same decoder serves every number of loops. Stages~1, 2 and~4 are those of TabICLv2.}
  \label{fig:backbone_stages}
\end{figure}

\paragraph{Input.} A table has $T=n_{\mathrm{tr}}+n_{\mathrm{q}}$ rows and $m$ feature columns, $X\in\mathbb{R}^{T\times m}$. The $n_{\mathrm{tr}}$ context rows come with labels $y_i$; the $n_{\mathrm{q}}$ query rows do not. Features are preprocessed as in TabICLv2 (Appendix~\ref{app:eval}). The model has no parameters specific to a particular column or dataset, so one set of weights serves tables of any width and length.

\paragraph{Stage 1: column-wise embedding.} The first stage gives every cell a $128$-dimensional embedding that reflects how its value sits within the distribution of its column. It has three ingredients.
\begin{itemize}[leftmargin=1.2em,itemsep=2pt,topsep=3pt,parsep=0pt]
\item \emph{Repeated feature grouping.} Instead of embedding each column on its own, column $j$ is embedded together with two others. A shared linear layer maps the triple of cell values to the embedding width,
$E^{(1)}_{ij}=\mathrm{Lin}\bigl(x_{i,j},\,x_{i,(j+1)\bmod m},\,x_{i,(j+3)\bmod m}\bigr)$,
so that every column appears in three groups and, for $m\geq7$, no two columns share more than one group.
\item \emph{Target-aware embedding.} The label of each context row is embedded (a lookup table for classification, a linear layer for regression) and added to all of that row's cell embeddings, $E^{(2)}_{ij}=E^{(1)}_{ij}+\mathrm{Emb}(y_i)$, so that the column encoder already sees how each feature relates to the target. Query rows receive no label embedding.
\item \emph{Induced set attention.} Each column is then processed as a set of $T$ cells by a set transformer with 3 blocks, 8 heads and 128 learned inducing points. In every block the inducing points first attend to the cells of the context rows and all cells then attend to the inducing points. Information therefore flows from context rows to all rows, never from query rows, and the cost is linear in the number of rows.
\end{itemize}

\paragraph{Scalable softmax.} Attention over many rows becomes diffuse, because a softmax spreads its mass more thinly as the number of keys $N$ grows. TabICLv2 counteracts this with a query-aware scalable softmax (QASSMax), which rescales each query before the dot product by a factor that depends on $\log N$ and on the query itself:
\begin{equation}
\tilde q_{h} \;=\; q_{h}\odot \mathrm{MLP}_{\mathrm{base}}(\log N)_h \odot \bigl(1+\tanh(\mathrm{MLP}_{\mathrm{gate}}(q_h))\bigr),
\end{equation}
for head $h$, where both MLPs have two layers and 64 hidden units. It is used in the inducing-point attention of the column stage and in every ICL block, and it is what lets a model trained on tables of 1{,}024 rows remain sharp on much longer ones.

\paragraph{Stage 2: row-wise interaction.} The second stage compresses each row into a single vector. For every row, the $m$ cell embeddings are prefixed with 4 learned [CLS] tokens and passed through a transformer with 3 blocks, width 128 and 8 heads, with rotary position embeddings (RoPE, base $10^5$) over the feature positions. The outputs of the 4 [CLS] tokens are concatenated into one $512$-dimensional row embedding. Attention here runs across the features of one row, so the cost is quadratic in the number of features and linear in the number of rows.

\paragraph{Stage 3: in-context learning.} Stages 1 and 2 leave one $512$-dimensional vector per row. A learned embedding of the label is added to the vectors of the context rows, which gives the ICL input $X_0\in\mathbb{R}^{T\times 512}$. The ICL stage is a transformer over the rows of the table: context rows attend only to context rows, and query rows attend only to context rows, so predictions for different query rows are independent and the context acts as the training set. Each block is a pre-norm transformer block with 8 heads, an MLP ratio of 2 and QASSMax attention (Appendix~\ref{app:icl_block}). In TabICLv2 this stage has 12 blocks with separate parameters and holds 25.7M of the model's 27.6M parameters, which is why it is the stage we study.

\paragraph{Stage 4: decoding.} A final LayerNorm and a small head map the ICL output of each query row to a prediction. The classifier outputs logits over 10 classes; tables with more classes are handled as in TabICLv2, by decomposing the label into digits of at most 10 values each and combining the predictions for the digits. The regressor predicts 999 quantiles of the target, at levels $0.001,\dots,0.999$, and is trained with the pinball loss; its point prediction is the mean of the quantiles. Classification and regression use separate models.

\paragraph{What \method{} changes.} Stages 1, 2 and 4 are used exactly as released, and run once per forward pass. \method{} replaces the 12 untied blocks of stage 3 with 4 weight-shared blocks applied for $k$ loops, and adds the cached stage input $X_0$ back before every block application after the first (Equation~\ref{eq:looped_icl}). Because stages 1 and 2 are not repeated, the extra cost of a loop is only that of the ICL blocks, $4$ block applications per loop, each quadratic in the number of context rows. The control replaces the 12 blocks with 4 untied blocks applied once, and has no injection.

\begin{table}[!ht]
  \centering\small
  \caption{\textbf{Notation used in the paper.}}
  \label{tab:notation}
  \begin{tabular}{@{}ll@{}}
    \toprule
    \textbf{Symbol} & \textbf{Meaning} \\
    \midrule
    $T = n_{\mathrm{tr}} + n_{\mathrm{q}}$ & rows in a table: labeled context rows plus query rows \\
    $d_{\mathrm{model}}$ & ICL width (512) \\
    $X_0 \in \mathbb{R}^{T\times d_{\mathrm{model}}}$ & ICL stage input: row embeddings plus context-label embeddings \\
    $n$ & number of distinct ICL blocks (12 in the baseline, 4 in \method{} and the control) \\
    $k$ & loop count; one loop applies each of the $n$ blocks once \\
    $d = nk$ & depth: the number of ICL block applications \\
    \shape{n}{k} & a model with $n$ distinct blocks applied $k$ times; \shape{n}{1} is an untied stack \\
    $B_\ell$, $C_j$ & untied block $\ell$; shared \method{} block $j$ \\
    $S_i$ & ICL state after $i$ block applications ($S_0 = X_0$) \\
    $g_\theta$, $p^{(k)}_{\theta,t}$ & decoder; predicted class distribution for query row $t$ after $k$ loops \\
    $\ell_{\mathrm{CE}}$ & training loss: query cross-entropy at the sampled loop count \\
    $L$ & held-out log loss on the 1{,}024 held-out synthetic tables \\
    \bottomrule
  \end{tabular}
\end{table}

\subsection{Summary of Components and ICL Variants}
\label{app:arch}

All models use the TabICLv2 backbone~\citep{qu2026tabiclv2} and change only its in-context-learning (ICL) stage. The column encoder, row encoder, label embedding and decoder are inherited without modification and run once per forward pass; their output $X_0\in\mathbb{R}^{T\times d_{\mathrm{model}}}$ is cached and reused as the injected input at every block application (Eq.~\ref{eq:looped_icl}). Table~\ref{tab:app-arch} lists the components.

\begin{table}[!ht]
  \centering\small
  \caption{\textbf{Backbone components.} Shared by all three models; only the ICL stage differs.}
  \label{tab:app-arch}
  \begin{tabular}{@{}llp{8.2cm}@{}}
    \toprule
    \textbf{Stage} & \textbf{Blocks / width} & \textbf{Details} \\
    \midrule
    Column & 3 blocks, width 128, 8 heads & Induced set attention, 128 inducing points, target- and query-aware scalable softmax; applied to groups of 3 features. \\
    Row    & 3 blocks, width 128, 8 heads & RoPE with base $10^5$; 4 CLS tokens are concatenated into a 512-dimensional row embedding. \\
    ICL    & width 512, 8 heads           & Pre-norm transformer blocks, MLP ratio 2, GELU, no dropout, query-aware scalable softmax. Test rows attend to context rows; context rows attend only to each other. \\
    Head   & --                           & Classifier: logits over 10 classes. Regressor: bias-free LayerNorm, linear target embeddings and a 999-quantile head (as released). \\
    \bottomrule
  \end{tabular}
\end{table}

\noindent The three ICL stages compared in the paper are:
\begin{itemize}
  \item \emph{baseline}: 12 distinct blocks, each applied once (the released architecture).
  \item \emph{\method{} (looped)}: 4 weight-shared blocks applied in sequence $k$ times, giving $4k$ block applications. The stage input $h_0$ is added to the residual stream before every application after the first ($x \leftarrow x + h_0$); the injection has no parameters.
  \item \emph{control}: the same 4 blocks applied once, untied and without injection.
\end{itemize}
Looped and control have identical parameter counts, and at $k=3$ the looped model performs the baseline's 12 block applications.

\subsection{Parameter Counts}
\label{app:params}

Each ICL block has 2.145M parameters, so the 12-block and 4-block ICL stacks hold $25.74$M and $8.58$M in \emph{blocks}. The ICL stage column of Table~\ref{tab:app-params} includes a further 0.54M outside the blocks (final LayerNorm, class-label embedding, and output projection); the main text quotes block-only counts for the ICL stage and uses stage totals elsewhere.

\begin{table}[!ht]
  \centering\small
  \caption{\textbf{Parameter counts (millions).} Column and row stages are identical across ICL variants.}
  \label{tab:app-params}
  \begin{tabular}{@{}lccccc@{}}
    \toprule
    & \textbf{Column} & \textbf{Row} & \textbf{ICL stage} & \textbf{Total} & \textbf{ICL applications} \\
    \midrule
    Classifier, baseline (12 untied)    & 0.88 & 0.40 & 26.28 & \textbf{27.55} & 12 \\
    Classifier, \method{} (4 shared)       & 0.88 & 0.40 &  9.12 & \textbf{10.40} & $4k$ \\
    Classifier, control (4 untied)      & 0.88 & 0.40 &  9.12 & \textbf{10.40} & 4 \\
    \midrule
    Regressor, baseline (12 untied)     & \multicolumn{3}{c}{--} & \textbf{28.54} & 12 \\
    Regressor, \method{} / control       & \multicolumn{3}{c}{--} & \textbf{11.40} & $4k$ / 4 \\
    \bottomrule
  \end{tabular}
\end{table}

\subsection{ICL Block}
\label{app:icl_block}

Each ICL block $C_j$ (and each baseline block $B_\ell$) is a standard pre-norm transformer block:
\begin{align}
    Z        &= S + \operatorname{MHA}\big(\operatorname{LN}_1(S);\, M\big), \label{eq:block-mha}\\
    C_j(S)   &= Z + \operatorname{FFN}\big(\operatorname{LN}_2(Z)\big),       \label{eq:block-ffn}
\end{align}
where $M$ is the attention mask (context rows attend only to context rows; query rows attend to context rows only), $\operatorname{MHA}$ is multi-head self-attention with query-aware scalable softmax, and $\operatorname{FFN}$ is a two-layer GELU MLP. Table~\ref{tab:icl_hparams} lists the hyperparameters; the block is identical in all models.

\begin{table}[!ht]
\centering\small
\caption{\textbf{ICL block hyperparameters.} Identical for all three models.}
\label{tab:icl_hparams}
\begin{tabular}{@{}lr@{}}
\toprule
Model width $d$                                & 512 \\
Attention heads / head dimension               & 8 / 64 \\
FFN hidden width (MLP ratio 2)                 & 1024 \\
Activation                                     & GELU \\
Normalization                                  & Pre-LayerNorm \\
Dropout                                        & 0.0 \\
Parameters per block                           & $\approx 2.14$M \\
\midrule
Distinct ICL blocks (baseline / \method{})      & 12 / 4 \\
Block parameters (baseline / \method{})         & 25.7M / 8.6M \\
Total parameters (baseline / \method{})         & 27.6M / 10.4M \\
\bottomrule
\end{tabular}
\end{table}

\begin{figure}[!htbp]
  \centering
  \includegraphics[width=0.85\linewidth]{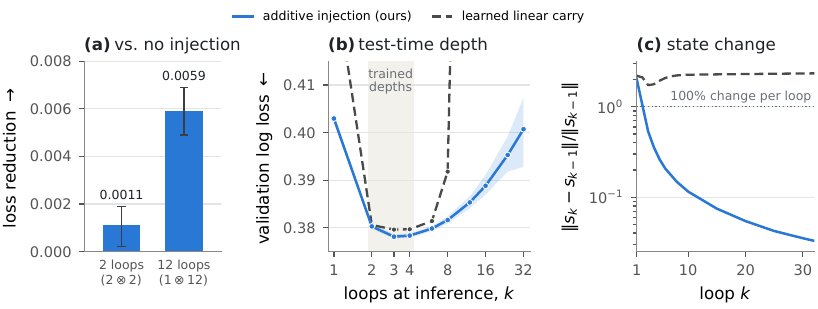}
  \caption{\textbf{Input injection} (small-scale study: width 128, shared ICL blocks, $10^{17}$ training FLOPs per run). \textbf{(a)}~Reduction in validation log loss from re-adding $X_0$ before every block application, against no injection, at 2 loops (shared $2\times2$) and 12 loops (shared $1\times12$), with paired 95\% intervals. \textbf{(b)}~Validation log loss against test-time depth for additive injection and the learned linear carry, both trained with $k\sim\mathcal{U}\{2,3,4\}$; mean and range over two seeds. The learned carry reaches 1.14 at $k=32$, off the axis. \textbf{(c)}~Relative change of the ICL state per loop, median over 512 validation tables. It shrinks steadily with additive injection but stays above 100\% per loop with the learned carry, whose loss then diverges beyond the trained depths.}
  \label{fig:injection}
\end{figure}

\subsection{Input Injection}
\label{app:injection}
Before every block application $i>0$, the cached stage input $X_0$ is added to the current state, so the block input is $S_i + X_0$. The injection is applied uniformly (also between blocks within a loop) and has no learned parameters or scaling. We compared it with two alternatives in a small-scale study (Figure~\ref{fig:injection}):

\begin{itemize}
    \item \emph{No injection:} blocks receive only $S_i$. Re-adding $X_0$ lowers validation log loss by 0.0011 nats at 2 loops and by 0.0059 nats at 12 loops; its benefit grows with depth.
    \item \emph{Learned linear carry:} $S_i \leftarrow W[S_i; X_0]$ with $W\in\mathbb{R}^{d\times 2d}$ initialized to $[I,I]$. Slightly worse than additive injection at the trained depths, and left unconstrained the carry becomes expansive: the largest singular value of its state block grows to 2.2, and the state changes by more than its own norm per loop. Its loss diverges beyond the trained range, reaching $1.14$ nats at $k=32$ against about $0.40$ with additive injection.
\end{itemize}
We therefore use the parameter-free additive injection and leave constrained carries to future work.

\clearpage
\section{Pretraining}
\label{app:pretraining}

\emph{Supports Sections~\ref{sec:training} and~\ref{sec:design}.} This appendix describes the synthetic data prior (Appendix~\ref{app:prior}), the three-stage curriculum with the loop-count schedules (Appendix~\ref{app:setup}), and the optimizer and hyperparameters (Appendix~\ref{app:hparams}). All models are pretrained only on synthetic tables.

\subsection{Synthetic Data Prior}
\label{app:prior}

All models are trained on synthetic tables from the TabICLv2 prior (\texttt{graph\_scm} in \texttt{tabicl})~\citep{qu2026tabiclv2}, with the released default settings and both filters enabled.

\paragraph{Generating one table.}
The generator first samples a random directed acyclic graph, then assigns features and the target to nodes and evaluates the graph. Root nodes draw random points. Every other node aggregates its parents through a random multi-function whose components are drawn from Gaussian-process (random Fourier features), generalized-GP, random-MLP, random oblivious-tree-forest, product, discretization, linear and normalizing-flow families. Categorical features (at most 200 categories) are produced by one of several discretization modes. Features are then clipped at 4 standard deviations, standardized and permuted. For classification the target node is discretized into $C$ classes and the labels are permuted; for regression it is kept real-valued.

\paragraph{Rejection filters.}
A table is regenerated if any of the following holds:
\begin{itemize}
  \item a feature is constant;
  \item for classification, the context and test splits do not contain the same set of classes, or fewer than two classes appear;
  \item an out-of-bag ExtraTrees regressor (25 trees, depth 6) does not beat the constant predictor at bootstrap $p<0.05$ (\emph{unpredictable-dataset filter});
  \item feature and target nodes share no common ancestor in the graph (\emph{unpredictable-graph filter}).
\end{itemize}

\paragraph{Table size and batching.}
Tables are generated in groups. All tables in a group share the number of rows and the context fraction; all tables in a subgroup share the number of features, $d\sim\mathcal{U}[d_{\min},d_{\max}(n)]$. The number of classes is drawn per table, $C\sim\mathcal{U}\{2,\dots,10\}$. The feature cap $d_{\max}(n)$ is 100 for $n\le10{,}240$ rows and falls with table length: 80, 60, 40, 30 and 20 features for at most 20K, 30K, 40K, 50K and 60K rows. Table~\ref{tab:app-prior} gives the per-stage settings. A group is exactly one training micro-batch in stage~1; in stages~2--3, micro-batches are packed from consecutive tables of equal length (Appendix~\ref{app:setup}).

\begin{table}[!ht]
  \centering\small
  \caption{\textbf{Prior settings by stage.} Classification follows the replication trainer; regression follows the released regression scripts.}
  \label{tab:app-prior}
  \begin{tabular}{@{}llcccc@{}}
    \toprule
    \textbf{Task} & \textbf{Stage} & \textbf{Rows per table} & \textbf{Context fraction} & \textbf{Features} & \textbf{Tables/step} \\
    \midrule
    Classification & 1 & 1{,}024                     & $\mathcal{U}[0.3,0.9]$ per group & 2--100            & 64 \\
    Classification & 2 & 400--10{,}240 (log-uniform) & 0.80                              & 2--100            & 64 \\
    Classification & 3 & 400--60{,}000 (log-uniform) & 0.80                              & 2--$d_{\max}(n)$  & 64 \\
    \midrule
    Regression     & 1 & 1{,}024                     & $\mathcal{U}[0.3,0.9]$ per group & 1--100            & 64 \\
    Regression     & 2 & 400--10{,}240 (log-uniform) & $\mathcal{U}[0.79,0.81]$          & 1--100            & 64 \\
    Regression     & 3 & 400--60{,}000 (log-uniform) & $\mathcal{U}[0.79,0.81]$          & 1--$d_{\max}(n)$  & 64 \\
    \bottomrule
  \end{tabular}
\end{table}

\paragraph{Streams and held-out set.}
Tables are generated on CPU nodes while the GPUs train. Classification streams come from a single 64-core node that serves the batches of step $i$ in order (seed 42 for the training streams). This stream is bit-exactly replayable: a resumed run receives exactly the batches it would have seen. Regression streams use seed 43 and are produced by pull-based worker pools on heterogeneous cluster nodes, so they are not bit-reproducible across machines. The held-out prior sets used for the loss curves are separate draws with seed 4242 (16 steps of 64 tables); they are generated once, cached and reused for every checkpoint.

\subsection{Curriculum and Depth Schedules}
\label{app:setup}

\paragraph{Three stages.}
Following TabICLv2, every model is pretrained in three stages. Stage~1 trains from scratch on short tables (500K steps). Stage~2 extends to 10K rows (40K steps). Stage~3 extends to 60K rows (10K steps). Each later stage is initialized from the previous stage's checkpoint with a fresh optimizer state. Stage~2 starts every model, the control included, from its 500K-step stage-1 checkpoint, so full training is data matched; the control's longer 660K-step run is used only for the compute-matched stage-1 comparison. The loss is cross-entropy over the test rows for classification; for regression it is the pinball loss averaged over 999 quantile levels, on targets standardized with the context rows' mean and standard deviation. Losses are averaged with equal weight per table.

% \paragraph{Codebases and replication.}
% The classifiers are trained with the single-GPU replication trainer of TabICLv2 stage~1. Its baseline reaches TabArena-Lite Elo $1538$ against $1558$ for the released classifier on the same harness. The ICL stage is replaced by a runtime patch to the encoder, and nothing else in the model or trainer changes. The regressors are trained with our package, TabLoop, configured to the released regression scripts. With identical weights it reproduces the official regressor's predictions to within $2\times10^{-4}$ in quantile space, and a replication of the official regressor trained this way reaches the released model's error on TabArena. Because the two tasks come from different replications, their recipes differ (Table~\ref{tab:app-hparams}). The comparisons in the paper are always within one task, between models trained with the same recipe.

\paragraph{Depth schedules for \method{}.}
The loop count $k$ is drawn independently for every micro-batch, with distribution depending on task and stage:
\begin{itemize}
  \item \emph{Stage 1, both tasks:} $k\sim\mathcal{U}\{2,3,4\}$ (mean 3). Expected training compute per step then equals the baseline's. Gradients flow through all $4k$ applications.
  \item \emph{Stages 2--3, regression:} the stage-1 schedule $\mathcal{U}\{2,3,4\}$, without truncation.
  \item \emph{Stages 2--3, classification:} four schedules, all continuing the same 500K-step stage-1 checkpoint on the same tables:
    \begin{itemize}
      \item \emph{Stage-1 schedule:} $\mathcal{U}\{2,3,4\}$ (mean 3), the baseline's depth in expectation;
      \item \emph{Uniform:} $\mathcal{U}\{2,\dots,8\}$ (mean 5);
      \item \emph{Truncated Poisson:} $k-2\sim\mathrm{Poisson}(3.205)$ renormalized to $\{2,\dots,8\}$ (mean 5);
      \item \emph{Spread-out:} $k=\min(2+\mathrm{Poisson}(10),32)$ (mean 12), with a latent-consistency loss ($\lambda=1$).
    \end{itemize}
\end{itemize}
The latent-consistency term runs one extra loop and penalizes the relative change of the test-row states, $\lambda\,\lVert s_{k+1}-\operatorname{sg}(s_k)\rVert^2 / \lVert \operatorname{sg}(s_k)\rVert^2$. In classification stages~2--3 we backpropagate through the last $\min(8,k+1)$ loops; earlier loops run without gradient, and their output is detached. The control is trained at $k=1$ in every stage. Each saved \method{} checkpoint uses the mean of its training schedule as its default inference depth; every result in the paper states $k$ explicitly instead of relying on that default.

\begin{table}[!htbp]
\centering
\footnotesize
\caption{\textbf{Models compared.} Stage~1 is matched in measured compute (18.2~EFLOP per model); the control's 24\% cheaper steps buy it 660K steps on 32\% more tables. Full training is data matched: every model, the control included, continues from its 500K-step stage-1 checkpoint through stages~2 and~3 (40K steps on up to 10{,}240 rows, then 10K on up to 60{,}000) on the same tables. Swatches use the colors of Figure~\ref{fig:schedule}.}
\label{tab:models}\label{tab:models_stage23}
\setlength{\tabcolsep}{3.5pt}\renewcommand{\arraystretch}{0.92}
\begin{tabular}{@{}lcccccc@{}}
\toprule
Model & Shape & Params & Appl. & Loop schedule & Steps & TFLOP/step \\
\midrule
\multicolumn{7}{@{}l}{\emph{Stage 1: compute matched}} \\
\swatch{baselinecol}\ Untied baseline & \shape{12}{1} & 27.6M & 12 & -- & 500K & 36.4 \\
\swatch{loopcol}\ \method{} & \shape{4}{k} & 10.4M & $4k$ & $\mathcal{U}\{2,3,4\}$ & 500K & 36.4 \\
\swatch{controlcol}\ Untied control & \shape{4}{1} & 10.4M & 4 & -- & 660K & 27.7 \\
\midrule
\multicolumn{7}{@{}l}{\emph{Stages 2 and 3: data matched, continuing each model's 500K-step stage-1 checkpoint}} \\
\swatch{baselinecol}\ Untied baseline & \shape{12}{1} & 27.6M & 12 & -- & 40K / 10K & -- \\
\swatch{controlcol}\ Untied control & \shape{4}{1} & 10.4M & 4 & -- & 40K / 10K & -- \\
\swatch{loopcol}\ \method{}, stage-1 schedule & \shape{4}{k} & 10.4M & $4k$ & $\mathcal{U}\{2,3,4\}$ & 40K / 10K & -- \\
\swatch{uniformcol}\ \method{}, uniform & \shape{4}{k} & 10.4M & $4k$ & $\mathcal{U}\{2,\dots,8\}$ & 40K / 10K & -- \\
\swatch{truncpoiscol}\ \method{}, trunc.\ Pois. & \shape{4}{k} & 10.4M & $4k$ & trunc.\ Poisson, 2--8 & 40K / 10K & -- \\
\swatch{spreadcol}\ \method{}, spread-out & \shape{4}{k} & 10.4M & $4k$ & $\min\{2{+}\mathrm{Pois}(10),32\}$ & 40K / 10K & -- \\
\bottomrule
\end{tabular}
\end{table}

Table~\ref{tab:models} summarizes the classification models in all three stages.

\paragraph{Micro-batching.}
In stage~1 a micro-batch is one generator group: 8 tables for the classification baseline and 16 for the looped and control classifiers. Gradient accumulation over micro-batches gives one update per 64 tables. The regressors use 4-table micro-batches. In stages~2--3, consecutive tables of equal length are packed into micro-batches of at most 20{,}000 rows (26{,}000 for the classification baseline; regression uses one table per micro-batch). Each micro-batch's loss is weighted by its share of the step's 64 tables, so packing changes memory, not the gradient. In stage~3, tables longer than 20{,}000 rows recompute activations in the backward pass; this is exact because dropout is zero.

\paragraph{Checkpoints and monitoring.}
Checkpoints are written every 10{,}000 steps in stage~1, every 1{,}000 steps in stage~2 and every 500 steps in stage~3. A watchdog restarts a dead trainer from its latest checkpoint, and halts the run instead if the loss becomes non-finite, if the average loss stays above $0.9$ after 12{,}000 steps, or if the scalable-softmax multiplier of the first ICL block exceeds 30 while the loss is high. No halt fired in the runs reported here. The classification trainer also skips any optimizer step whose gradient norm or loss is non-finite; no such step occurred in the baseline, looped or control stage-1 runs.

\subsection{Optimizer and Hyperparameters}
\label{app:hparams}

Both tasks use Muon~\citep{jordan2024muon} for the hidden weight matrices: Nesterov momentum, five Newton--Schulz iterations, and update scaled by $0.2\sqrt{\max(m,n)}$ to match AdamW's RMS~\citep{liu2025muon}. Classification additionally places embeddings, norms, heads, biases, scalable-softmax modulators and CLS tokens in an AdamW group; regression uses Muon on all parameters. The learning rate follows a linear warmup and a cosine decay. Weight sharing needs no change to the optimizer: a weight-shared block's gradient is the sum of the gradients of its $k$ applications, and Muon orthogonalizes it before the update, so the update size does not grow with $k$. All three ICL variants use the same learning rates (Table~\ref{tab:app-hparams}).

\begin{table}[!ht]
  \centering\footnotesize
  \caption{\textbf{Pretraining hyperparameters.} Values shared across stages of a task are listed once.}
  \label{tab:app-hparams}\label{tab:optim}
  \setlength{\tabcolsep}{3.5pt}\renewcommand{\arraystretch}{0.92}
  \begin{tabular}{@{}lccc|ccc@{}}
    \toprule
    & \multicolumn{3}{c|}{\textbf{Classification}} & \multicolumn{3}{c}{\textbf{Regression}} \\
    & stage 1 & stage 2 & stage 3 & stage 1 & stage 2 & stage 3 \\
    \midrule
    Steps                                    & 500K$^{a}$ & 40K & 10K & 500K$^{a}$ & 40K & 10K \\
    Tables per step                          & 64 & 64 & 64 & 64 & 64 & 64 \\
    Rows per table                           & 1{,}024 & $\le$10{,}240 & $\le$60{,}000 & 1{,}024 & $\le$10{,}240 & $\le$60{,}000 \\
    Optimizer                                & \multicolumn{3}{c|}{Muon + AdamW} & \multicolumn{3}{c}{Muon (all parameters)} \\
    Peak LR (Muon)                           & $8{\times}10^{-4}$ & $1{\times}10^{-4}$ & $2{\times}10^{-5}$ & $8{\times}10^{-4}$ & $1{\times}10^{-4}$ & $2{\times}10^{-5}$ \\
    Peak LR (AdamW)                          & $1{\times}10^{-4}$ & $1{\times}10^{-4}$ & $2{\times}10^{-5}$ & -- & -- & -- \\
    Muon momentum                            & \multicolumn{3}{c|}{0.95} & \multicolumn{3}{c}{0.9} \\
    AdamW $(\beta_1,\beta_2)$, $\epsilon$    & \multicolumn{3}{c|}{$(0.95, 0.95)$, $10^{-8}$} & \multicolumn{3}{c}{--} \\
    Weight decay (decoupled)                 & \multicolumn{3}{c|}{0.1} & \multicolumn{3}{c}{0.01} \\
    Warmup steps (linear)                    & 10{,}000$^{a}$ & 800 & 200 & 5{,}000$^{a}$ & 400 & 100 \\
    Final LR                                 & \multicolumn{3}{c|}{0} & \multicolumn{3}{c}{$10^{-7}$} \\
    Gradient clipping (global norm)          & 10 & 10 & 1 & 10 & 10 & 1 \\
    Precision                                & \multicolumn{3}{c|}{bf16 autocast, fp32 islands$^{b}$, TF32} & fp32, TF32 & \multicolumn{2}{c}{fp16 attention$^{c}$} \\
    Micro-batch (tables / rows)              & 8 / 16$^{d}$ & \multicolumn{2}{c|}{$\le$20K rows$^{e}$} & 4 tables & \multicolumn{2}{c}{1 table} \\
    \texttt{torch.compile}                   & per block & no & no & yes & yes & yes \\
    Seed of the reported runs (model, data)  & \multicolumn{3}{c|}{42, 42} & \multicolumn{3}{c}{43, 43} \\
    Depth $k$, \method{}                      & $\mathcal{U}\{2,3,4\}$ & \multicolumn{2}{c|}{see Appendix~\ref{app:setup}} & \multicolumn{3}{c}{$\mathcal{U}\{2,3,4\}$} \\
    \bottomrule
  \end{tabular}

  \vspace{1pt}
  \raggedright\footnotesize
  $^{a}$Controls: 660K steps with warmup 10{,}000 (classification) and 685K steps with warmup 6{,}850 (regression, 1\% of its steps). \quad
  $^{b}$Scalable-softmax modulators and all column-stage attention run in fp32; the rest in bf16. \quad
  $^{c}$Attention on fp16 queries, keys and values (the released recipe's FlashAttention-3 numerics), with dynamic loss scaling; the rest in fp32. \quad
  $^{d}$8 tables for the baseline, 16 for \method{} and the control. \quad
  $^{e}$26{,}000 rows for the baseline.
\end{table}

\paragraph{Deviations from the released recipes.}
Our classification recipe differs from the released TabICLv2 scripts in eight details: bf16 autocast with fp32 islands (not TF32); plain weight decay 0.1 (not cautious 0.01); a separate AdamW group for embeddings, norms, heads and scalable-softmax modulators (not a single Muon group); 2\% warmup (not 1\%); final learning rate 0 (not $10^{-7}$); at least two features per table (not one); micro-batches of 8 tables (not 4); and a context fraction of exactly $0.80$ in stages~2--3 (not $[0.79,0.81]$). Regression follows the released scripts, plus dynamic loss scaling for the fp16 attention of stages~2--3.

\clearpage
\section{Compute Accounting}
\label{app:compute_full}

\emph{Supports Section~\ref{sec:design}.} This appendix accounts for compute: the per-block FLOP cost and the inference cost of each configuration (Appendix~\ref{app:compute}) and the measured training compute and how it is matched (Appendix~\ref{app:train_compute}).

\subsection{Inference Compute}
\label{app:compute}

Per-application FLOPs are identical across models, since the block is identical. For a table with $T$ rows, one forward block application costs approximately $16\,T d^2 + 4\,T^2 d$ FLOPs (attention projections and FFN with MLP ratio 2, plus attention scores and mixing). We report inference compute in ICL block applications; the encoders and decoder are identical across models and excluded.

\begin{table}[!ht]
\centering\small
\caption{\textbf{Inference compute of the configurations in the main text}, counted in ICL block applications.}
\label{tab:compute}
\begin{tabular}{@{}lrr@{}}
\toprule
\textbf{Configuration} & \textbf{Block applications} & \textbf{\% of baseline} \\
\midrule
Untied baseline \shape{12}{1}                                  & 12  & 100 \\
Untied control \shape{4}{1}                                    & 4   & 33  \\
\method{}, $k=1$                                             & 4   & 33  \\
\method{}, $k=2$                                             & 8   & 67  \\
\method{}, $k=3$ (stage-1 default; equal to baseline)        & 12  & 100 \\
\method{}, $k=4$                                             & 16  & 133 \\
\method{}, $k=5$                                             & 20  & 167 \\
\method{}, $k=8$ (peak of the concentrated schedules)        & 32  & 267 \\
\method{}, $k=12$ (mean of the spread-out schedule)          & 48  & 400 \\
\method{}, $k=16$                                            & 64  & 533 \\
\method{}, $k=32$ (peak of the spread-out schedule)          & 128 & 1067 \\
\bottomrule
\end{tabular}
\end{table}

\subsection{Training Compute}
\label{app:train_compute}

Table~\ref{tab:models} in Section~\ref{sec:design} summarizes the three stage-1 classifiers, and Table~\ref{tab:app-flops} lists the stage-1 models of both tasks with their measured compute.

We measure training FLOPs per step with PyTorch's FLOP counter on replayed stage-1 batches, at the mean depth for the looped models. Stage-1 controls are trained longer to match the others' 500K steps in compute: 660K steps for the classifier and 685K steps for the regressor. The control classifier keeps the 10{,}000-step warmup; the control regressor's warmup stays at 1\% of its steps. Full training is data matched instead: stages~2 and~3 continue every model, the control included, from its 500K-step stage-1 checkpoint for the same steps on the same tables.

\begin{table}[!ht]
  \centering\small
  \caption{\textbf{Models of the full-scale comparison in stage~1}, in the notation \shape{n}{k} ($n$ distinct ICL blocks applied $k$ times). Applications per forward pass and TFLOP per step (forward and backward, measured) are at the mean sampled loop count. Each control's 500K-step checkpoint is the starting point of its data-matched full training. Totals: 54.7~EFLOP for the three classifiers and 46.6~EFLOP for the three regressors.}
  \label{tab:app-flops}
  \setlength{\tabcolsep}{4pt}
  \begin{tabular}{@{}lcccccccl@{}}
    \toprule
    Model & Blocks & Params & Loops sampled & Appl./fwd & TFLOP/step & Steps & EFLOP & Role \\
    \midrule
    \multicolumn{9}{@{}l}{\textit{Classification}} \\
    \shape{12}{1} & 12 untied & 27.6M & -- & 12 & 36.4 & 500K & 18.2 & untied baseline \\
    \shape{4}{3}  & 4 shared    & 10.4M & $\mathcal U\{2,3,4\}$ & 12 & 36.4 & 500K & 18.2 & \method{} \\
    \shape{4}{1}  & 4 untied  & 10.4M & -- & 4 & 27.7 & 660K & 18.2 & its control \\
    \midrule
    \multicolumn{9}{@{}l}{\textit{Regression}} \\
    \shape{12}{1} & 12 untied & 28.5M & -- & 12 & 31.1 & 500K & 15.5 & untied baseline \\
    \shape{4}{3}  & 4 shared    & 11.4M & $\mathcal U\{2,3,4\}$ & 12 & 31.1 & 500K & 15.5 & \method{} \\
    \shape{4}{1}  & 4 untied  & 11.4M & -- & 4 & 22.7 & 685K & 15.5 & its control \\
    \bottomrule
  \end{tabular}
\end{table}

\paragraph{Training compute in stages 2 and 3.}
Stage 1 is matched in measured FLOPs (Table~\ref{tab:models} and above). Full training is matched in data: every model continues from its 500K-step stage-1 checkpoint on the same tables. A baseline forward pass applies 12 blocks; a \method{} forward pass applies $4(\bar k+1)$ blocks, where $\bar k$ is the mean sampled loop count and the extra loop is a consistency loop that the stage-2/3 trainer runs beyond $k$ in every \method{} step, also when the consistency loss is off, and which costs compute but contributes no gradient: about 24 under the concentrated schedules ($\bar k=5$) and about 52 under the spread-out schedule ($\bar k\approx 12$). Backpropagation covers at most the last 8 loops. The control applies 4 blocks. Stage-2 and stage-3 gains of \method{} over the baseline are therefore not compute-matched; this is noted wherever those numbers appear (Section~\ref{sec:frontier}).

\clearpage
\section{Evaluation Protocols}
\label{app:eval}

\emph{Supports Section~\ref{sec:design}.} All models are applied zero-shot with TabICLv2's default input preprocessing. The four evaluation suites used in the main text are:

\begin{description}
    \item[Held-out prior.] Mean query log loss on a cached held-out slice of 1{,}024 tables (16 micro-batches of 64) drawn from the training prior with seed 4242, disjoint from training. The same slice is reused for every checkpoint of every model, so comparisons are paired. \method{} is evaluated at $k\in\{1,\dots,6\}$ at every checkpoint; the baseline and control at their fixed depth.
    \item[TabArena.] We follow the TabArena protocol~\citep{erickson2025tabarena} on its 38 classification datasets and all 594 outer-fold splits, and compute Elo against the public leaderboard with 95\% bootstrap intervals. Intermediate stage-1 checkpoints (Appendix~\ref{app:intermediate}) and the spread-out stage-2 model in Figure~\ref{fig:stage2} are also evaluated with TabArena-Lite (one split per dataset); its Elo is not comparable with full TabArena. Every Elo in the main text is full TabArena unless marked.
    \item[TALENT.] We evaluate on the 200 classification datasets of TALENT~\citep{liu2024talenttabularanalyticslearning} with the official train/validation/test splits and report mean test accuracy.
    \item[Regression.] We evaluate the regressors after stage~3 on the 13 TabArena regression datasets, reporting Elo against the regression leaderboard, and on 100 TALENT regression datasets, reporting each model's win rate by RMSE against the other configurations (50\% is average). Regression Elo is not comparable with classification Elo.
\end{description}

\paragraph{Choice of loop count.}
\method{} is read at a fixed $k$ for each evaluation. Where a table reports the \emph{best} $k$, it is the maximizer over the evaluated depths on that benchmark itself (so slightly optimistic); the full curves are in Figure~\ref{fig:schedule}c, d. A fixed $k=8$ reproduces the headline TabArena result for the concentrated schedules.

\paragraph{Random seeds.}
We repeat the experiments with multiple random seeds, and the conclusions of the main text hold across them: \method{} read at three or more loops performs comparably to the untied baseline, both stay ahead of the untied control, and predictions improve with the number of loops within the trained range. Differences between \method{} and the baseline that are within the variation across seeds are read as ties throughout.

\clearpage
\apppart{partII}{Part II\texorpdfstring{\quad}{: }Results}
\section{Depth--Parameter Scaling Study}
\label{app:sweep}
\emph{Supports Section~\ref{sec:sweep}.} Figure~\ref{fig:sweep_analysis} in Section~\ref{sec:sweep} gives the main results of the depth--parameter scaling study. This appendix lists its models and the deeper pair (Appendix~\ref{app:sweep_models}), enlarges the synthetic panels and gives the fitted curves (Appendix~\ref{app:sweep_fits}), shows how the gain from looping varies with the block count, over training and across tables (Appendix~\ref{app:sweep_training}), and reports the full real-data results (Appendix~\ref{app:sweep_real}).

\subsection{Models of the Study}
\label{app:sweep_models}
The study trains eleven classifiers for 3.20~EFLOP each, at 32 tables per step, with training compute measured as in Appendix~\ref{app:train_compute} (Table~\ref{tab:sweep_models}). All use the ICL block of the full-scale models, so a model with $n$ distinct blocks has $1.82\text{M}+n\times2.14\text{M}$ parameters. The five looped models \shape{n}{12/n}, $n\in\{1,2,3,4,6\}$, sample their loop count around $12/n$ so that each makes twelve block applications per step on average, and train for 200K steps. The six untied models \shape{n}{1}, $n\in\{1,2,3,4,6,12\}$, apply their blocks once; with fewer than twelve blocks they cost less per step and therefore train for more steps at the same compute, and their 200K-step checkpoints give the equal-steps comparison of Figure~\ref{fig:sweep_analysis}a. As in every comparison (Section~\ref{sec:design}), the models share the prior, encoders, decoder, optimizer and schedule, and all are scored on the same 1{,}024 held-out synthetic tables (Appendix~\ref{app:eval}), so every difference in Figure~\ref{fig:sweep_analysis} is paired table by table. Figure~\ref{fig:sweep_analysis}c reads each looped model over the loop counts it was trained on. A deeper pair doubles the depth: \shape{4}{6}, which samples $k$ from $\{4,\dots,8\}$ and so makes 24 block applications per step on average, and 24 untied blocks \shape{24}{1}, both trained for 200K steps (6.4M tables) at 4.45~EFLOP each (Figure~\ref{fig:sweep_analysis}b). Two points qualify this pair. First, \shape{24}{1} is reported from a rerun with a gentler recipe (Muon learning rate $5.7\times10^{-4}$ instead of $8\times10^{-4}$, AdamW $7\times10^{-5}$ instead of $10^{-4}$, gradient clipping at 1 instead of 10), because with the shared recipe its loss spiked between 85K and 130K steps and it finished at 0.4756; the rerun reaches 0.4672. Second, \shape{4}{6} and \shape{24}{1} were trained on separately generated streams from the same prior, with the same number of tables but not the identical tables that the other models of the study share. \shape{4}{6} is best at $k=7$ and $k=8$, which both give 0.4663, and reaches 0.4665 at $k=6$, the mean of its training depths.

Figure~\ref{fig:sweep_analysis}b is scored on a second set of unseen tables from the training prior, on which all losses are about 0.015 lower than on the held-out slice used elsewhere; the losses quoted in this appendix refer to the original slice. That panel also includes a third looped model, \shape{4}{12}, which samples $k$ from $\{8,\dots,16\}$ and is trained on the same 6.4M tables at 6.95~EFLOP; its untied counterpart has 48 blocks (104.8M parameters) and reaches 0.447, slightly below the looped model's 0.4476 at 48 block applications. Neither is listed in Table~\ref{tab:sweep_models}.

\begin{table}[!htbp]
  \centering\small
  \caption{\textbf{Models of the depth--parameter scaling study} (3.20~EFLOP each, and 4.45~EFLOP for the deeper pair; 13 runs, 44.1~EFLOP in total; 32 tables per step). Applications per forward pass and TFLOP per step are at the mean sampled loop count; TFLOP per step is measured (forward and backward), except for one block, where it comes from the linear fit $9.77+0.52\times$ applications to the measured values. The looped models, \shape{12}{1} and \shape{24}{1} see 6.4M tables; the untied controls train longer and see 8.0M to 10.0M. \shape{6}{2} samples $k=1$, unlike the other looped models.}
  \label{tab:sweep_models}
  \setlength{\tabcolsep}{3.2pt}
  \begin{tabular}{@{}lcccccccl@{}}
    \toprule
    Model & Blocks & Params & Loops sampled & Appl./fwd & TFLOP/step & Steps & EFLOP & Role \\
    \midrule
    \shape{1}{12} & 1 shared    & 4.0M & $\mathcal U\{8,\dots,16\}$ & 12 & 16.0 & 200K & 3.2 & single block \\
    \shape{1}{1}  & 1 untied  & 4.0M & --  & 1  & 10.3 & 311K & 3.2 & its control \\
    \addlinespace[2pt]
    \shape{2}{6}  & 2 shared    & 6.1M  & $\mathcal U\{4,\dots,8\}$ & 12 & 16.0 & 200K & 3.2 & \\
    \shape{2}{1}  & 2 untied  & 6.1M  & --  & 2  & 10.8 & 296K & 3.2 & its control \\
    \addlinespace[2pt]
    \shape{3}{4}  & 3 shared    & 8.3M  & $\mathcal U\{3,4,5\}$ & 12 & 16.0 & 200K & 3.2 & \\
    \shape{3}{1}  & 3 untied  & 8.3M  & --  & 3  & 11.3 & 282.6K & 3.2 & its control \\
    \addlinespace[2pt]
    \shape{4}{3}  & 4 shared    & 10.4M & $\mathcal U\{2,3,4\}$ & 12 & 16.0 & 200K & 3.2 & as \method{} \\
    \shape{4}{1}  & 4 untied  & 10.4M & --  & 4  & 11.9 & 270K & 3.2 & its control \\
    \addlinespace[2pt]
    \shape{6}{2}  & 6 shared    & 14.7M & $\mathcal U\{1,2,3\}$ & 12 & 16.0 & 200K & 3.2 & coarsest loop \\
    \shape{6}{1}  & 6 untied  & 14.7M & --  & 6  & 12.9 & 248K & 3.2 & its control \\
    \midrule
    \shape{12}{1} & 12 untied & 27.6M & --  & 12 & 16.0 & 200K & 3.2 & reference \\
    \midrule
    \shape{4}{6}  & 4 shared    & 10.4M & $\mathcal U\{4,\dots,8\}$ & 24 & 22.3 & 200K & 4.45 & twice the depth \\
    \shape{24}{1} & 24 untied & 53.3M & --  & 24 & 22.3 & 200K & 4.45 & reference (rerun) \\
    \bottomrule
  \end{tabular}
\end{table}

\subsection{Fitted Curves}
\label{app:sweep_fits}
Figure~\ref{fig:sweep_depth_full} enlarges panels (a) and (c) of Figure~\ref{fig:sweep_analysis}, with an inset on 6 to 18 block applications where the looped readings lie.

\begin{figure}[!htbp]
  \centering
  \includegraphics[width=\linewidth]{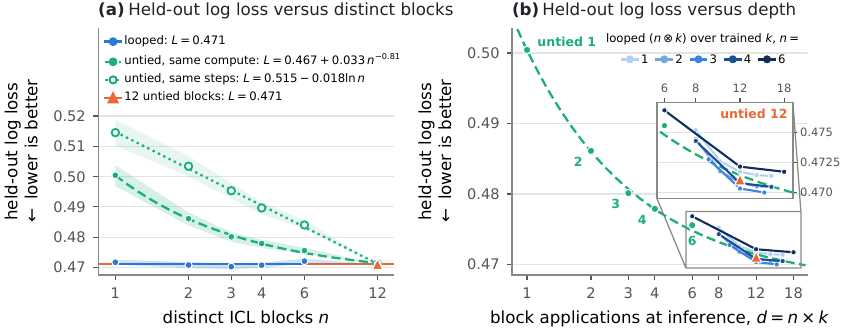}
  \caption{\textbf{Enlarged view of Figure~\ref{fig:sweep_analysis}a, c.} \textbf{(a)}~Held-out log loss against distinct blocks $n$, with the fits in the legend. \textbf{(b)}~Held-out log loss against block applications $d$; the inset enlarges 6 to 18 applications, where the looped models lie on the power law fitted to the untied models.}
  \label{fig:sweep_depth_full}
\end{figure}

The lines in Figure~\ref{fig:sweep_analysis}a and the dashed curve in Figure~\ref{fig:sweep_analysis}c are least-squares fits to the measured points; Table~\ref{tab:sweep_fits} gives their forms. Panel~(a) plots held-out loss itself, not its difference from the 12-block model, and the fits are made on that scale; the equal-compute series is the same in panels (a) and (c), where $n=d$, so the two panels share one fit. At equal steps a power law does not converge, because its exponent tends to zero, so we use its logarithmic limit. The fits in panel~(a) have one to three parameters for five or six points, and we use them as guides for the eye.

\textbf{A scaling law in depth.} Figure~\ref{fig:sweep_analysis}c tests whether loss depends on depth alone. The law $L=c+a\,d^{-b}$ is fitted to the six untied models only, and the panel plots the held-out log loss of every model against $d$ on a logarithmic depth axis, together with the fitted curve. The nineteen looped readings, which did not enter the fit, lie within 0.002 of it, with a root-mean-square error of 0.0010 and a mean offset of $+0.0004$. Fitting all 25 readings together gives nearly the same curve (Table~\ref{tab:sweep_fits}). Two cautions apply. First, the exponent is loosely determined: the floor $c$ and the exponent $b$ trade off against each other over one decade of depth, and a bootstrap over the 25 readings gives a 95\% interval of 0.20 to 0.88 for $b$ and 0.444 to 0.468 for $c$. What the data support is the collapse onto one curve, not a particular exponent. Second, the law describes models trained with 3.2~EFLOP; the deeper models of Figure~\ref{fig:sweep_analysis}b were trained with more compute and are scored on a different set of tables, so they are not placed on it.

\begin{table}[!htbp]
  \centering\small
  \caption{\textbf{Fitted curves of the depth--parameter scaling study.} $L$: held-out log loss; $n$: distinct blocks (equal to the number of block applications $d$ for untied models). The last column is the largest absolute difference between a fit and a measured point.}
  \label{tab:sweep_fits}
  \renewcommand{\arraystretch}{1.3}
  \setlength{\tabcolsep}{5pt}
  \begin{tabular}{@{}cllcc@{}}
    \toprule
    Panel & Series & Fitted form & Points & Max.\ residual \\
    \midrule
    (a) & \swatch{loopcol}\ Looped \shape{n}{12/n}            & $L = 0.4711$ (constant)                         & 5 & 0.0010 \\
    (a), (b) & \swatch{controlcol}\ Untied \shape{n}{1}, equal compute & $L = 0.467 + 0.033\,n^{-0.81}$ & 6 & 0.0008 \\
    (a) & \swatch{controlcol}\ Untied \shape{n}{1}, equal steps   & $L = 0.515 - 0.018\,\ln n$                      & 6 & 0.0008 \\
    (b) & All untied and looped readings, joint fit              & $L = 0.467 + 0.034\,d^{-0.76}$                  & 25 & 0.0016 \\
    \bottomrule
  \end{tabular}
\end{table}

\subsection{Over Training and Across Tables}
\label{app:sweep_training}
Figure~\ref{fig:sweep_price} gives the gain from looping for every block count, that is, how much better $n$ blocks do when they are looped to twelve applications than when they are applied once at the same compute, with the share of tables on which the looped model is better.

\begin{figure}[!htbp]
  \centering
  \includegraphics[width=0.65\linewidth]{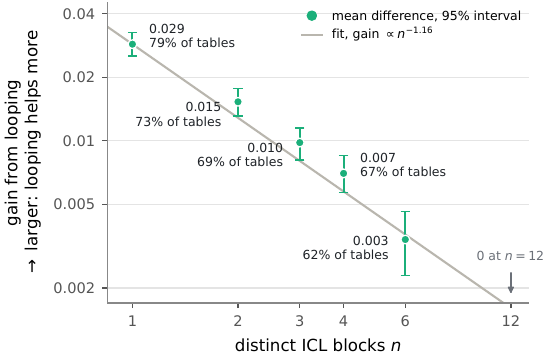}
  \caption{\textbf{Gain from looping at equal compute.} Held-out loss of \shape{n}{1} minus \shape{n}{12/n}, on a log scale with paired 95\% intervals. The two models have the same $n$ blocks and parameters and differ in depth; the effect of weight sharing at a fixed depth is the looped curve of Figure~\ref{fig:sweep_analysis}a. Points are labeled with the mean difference and the share of the 1{,}024 held-out tables on which the looped model is better. The line is a power-law fit, gap $\propto n^{-1.16}$.}
  \label{fig:sweep_price}
\end{figure}

For two to four blocks the advantage of looping over untying at the same parameter count appears early in training; for six blocks it emerges after about 1.3~EFLOP; all grow toward their final values (Figure~\ref{fig:sweep_compute}a). Intermediate points compare runs at different points of their own learning-rate schedules. Per table (Figure~\ref{fig:sweep_compute}b), two weight-shared blocks applied six times beat two untied blocks on 73\% of tables (mean difference 0.0152) and differ from twelve untied blocks by $-0.0003$ on average, being better on 54\% of tables.

\begin{figure}[!htbp]
  \centering
  \includegraphics[width=\linewidth]{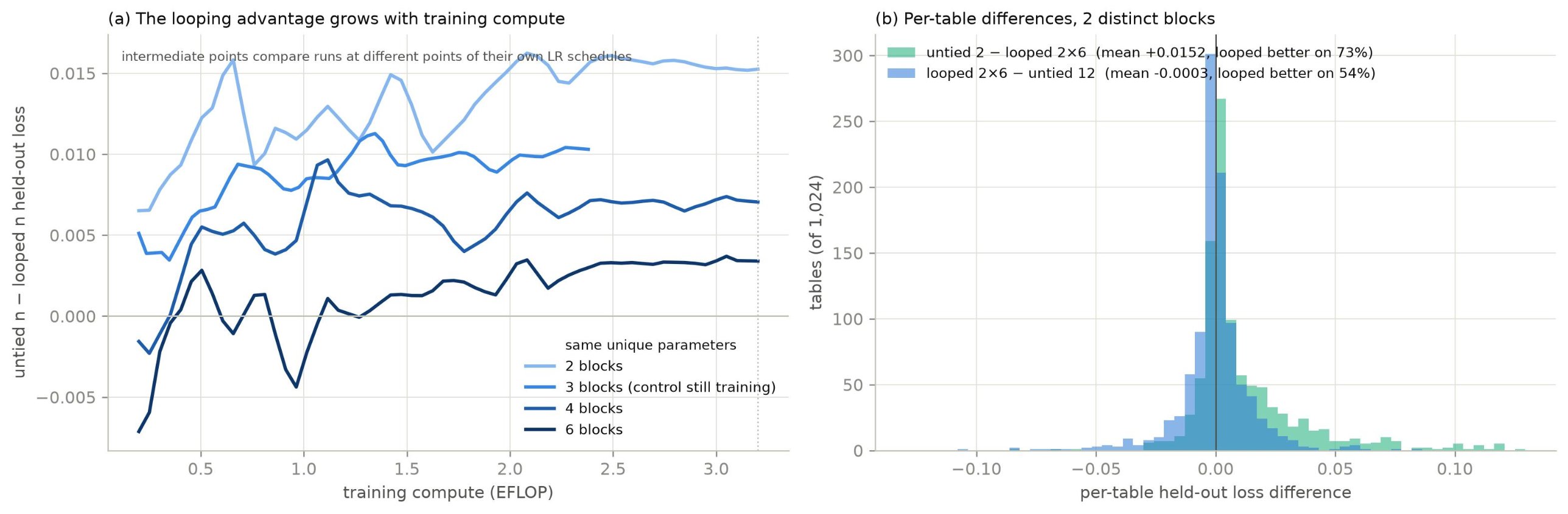}
  \caption{\textbf{Looping advantage over training and across tables.} \textbf{(a)}~Held-out loss of untied $n$ minus looped $n$ at the same parameter count, against training compute; intermediate points compare runs at different points of their own learning-rate schedules. This view was recorded before the three-block control finished (its curve stops at about 2.4~EFLOP) and before the one-block models were added; their final values are in Figure~\ref{fig:sweep_price}. \textbf{(b)}~Per-table held-out loss differences for two distinct blocks: two untied blocks minus two weight-shared blocks applied six times, and the looped model minus twelve untied blocks.}
  \label{fig:sweep_compute}
\end{figure}

\subsection{The Study on Real Data}
\label{app:sweep_real}
Figure~\ref{fig:sweep_analysis}e, f in Section~\ref{sec:sweep} summarizes these results; Figure~\ref{fig:sweep_real_full} evaluates the models of Figure~\ref{fig:sweep_analysis}a on the 38 classification datasets of TabArena-Lite: fifteen readings, the five looped models and the five untied models with fewer than twelve blocks at equal compute and at equal steps, against twelve untied blocks. The looped models stay close to twelve untied blocks in Elo and within about 3\% of them in per-dataset error. The untied models are 5 to 12\% worse at equal compute and 6 to 15\% worse at equal steps, and the gap grows as $n$ shrinks (panel~b). Compared directly with untied blocks of the same number, looping lowers the error on 21 to 29 of the 38 datasets for $n\le4$ at equal compute, and on 15 at $n=6$, where the difference is within its interval (panel~c). The real-data gap of each model follows its synthetic held-out gap, with a Pearson correlation of 0.95 over the fifteen readings (panel~d).

\begin{figure}[!htbp]
  \centering
  \includegraphics[width=\linewidth]{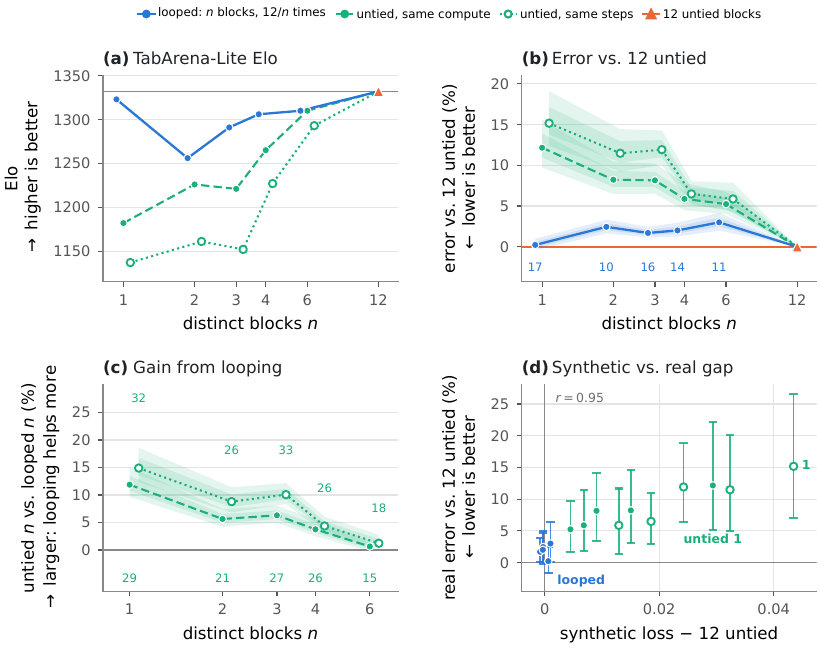}
  \caption{\textbf{The depth--parameter scaling study on real data} (TabArena-Lite, 38 classification datasets). \textbf{(a)}~Elo. \textbf{(b)}~Geometric mean of per-dataset error ratios against twelve untied blocks, shaded by the central 50\% paired interval; numbers count the datasets on which the looped model beats twelve untied blocks. \textbf{(c)}~Error of untied \shape{n}{1} relative to looped \shape{n}{12/n}, shaded by the central 50\% paired interval; numbers count the datasets on which the looped model is better, at equal steps (above) and at equal compute (below). \textbf{(d)}~The real-data gap of panel~(b) against the synthetic held-out gap of Figure~\ref{fig:sweep_analysis}a. Values were read from the evaluation plots.}
  \label{fig:sweep_real_full}
\end{figure}

\clearpage
\section{Full-Setup Comparison: Compute-Matched Stage~1 and Data-Matched Full Training}
\label{app:fullscale}
\label{app:regression}

\emph{Supports Section~\ref{sec:matched}.} This appendix covers both tasks of the full-scale comparison, side by side: the models (Appendix~\ref{app:reg_setup}), their held-out loss during stage~1 (Appendix~\ref{app:heldout}), the classifiers at two intermediate budgets of equal compute (Appendix~\ref{app:intermediate}), the regressors after data-matched full training (Appendix~\ref{app:reg_results}), and the differences between the regression runs.

\subsection{Models in Both Tasks}
\label{app:reg_setup}
The three classifiers are listed in Table~\ref{tab:models}. The regressors mirror the classifiers: an untied 12-block baseline (28.5M parameters), \method{} with 4 weight-shared blocks (11.4M), and an untied 4-block control (11.4M), all with a 999-quantile head trained with the pinball loss on the same synthetic prior. \method{} samples $k\sim\mathcal{U}\{2,3,4\}$ in every stage. Stage~1 is matched in compute (16~EFLOP: 500K steps for the baseline and \method{}, 685K for the control). Full training is data matched: every model continues from its 500K-step stage-1 checkpoint through stages~2 and~3 (40K and 10K steps) on the same tables, so the baseline and \method{} apply twelve blocks per step in expectation and the control applies four. Real-data results are reported after stage~3 on 13 TabArena and 100 TALENT regression datasets; these Elo values come from the regression leaderboard and are not comparable with the classification Elo in the main text.

\subsection{Held-out Loss During Stage 1}
\label{app:heldout}
\label{app:reg_heldout}
Panels (a, b) of Figure~\ref{fig:heldout} follow the query log loss of the three classifiers on 1{,}024 held-out synthetic tables through stage~1, against training steps and training compute, with \method{} read at one to four loops from the same checkpoints; the insets enlarge the end of training. The control trains for 660K cheaper steps, so that its total compute matches the 500K steps of the other runs. \method{} at $k=3$, which matches the baseline's 12 block applications, tracks the baseline throughout training; the control ends above every reading of \method{} with $k\ge2$; and $k=1$, a depth never trained, is far worse.

In regression, panels (c, d) follow the held-out pinball loss of the three regressors through stage~1, against training steps and training compute, with \method{} read at one to four loops from the same checkpoints. The control trains for 685K cheaper steps, so that its total compute matches the 500K steps of the other runs. As in classification, \method{} at $k=3$ (the baseline's 12 block applications) tracks the untied baseline throughout training, $k=4$ ends lowest, and the control ends highest among all readings with $k\ge2$; at $k=1$, a depth never trained, the loss is far higher.

\begin{figure}[!htbp]
  \centering
  \includegraphics[width=0.85\linewidth]{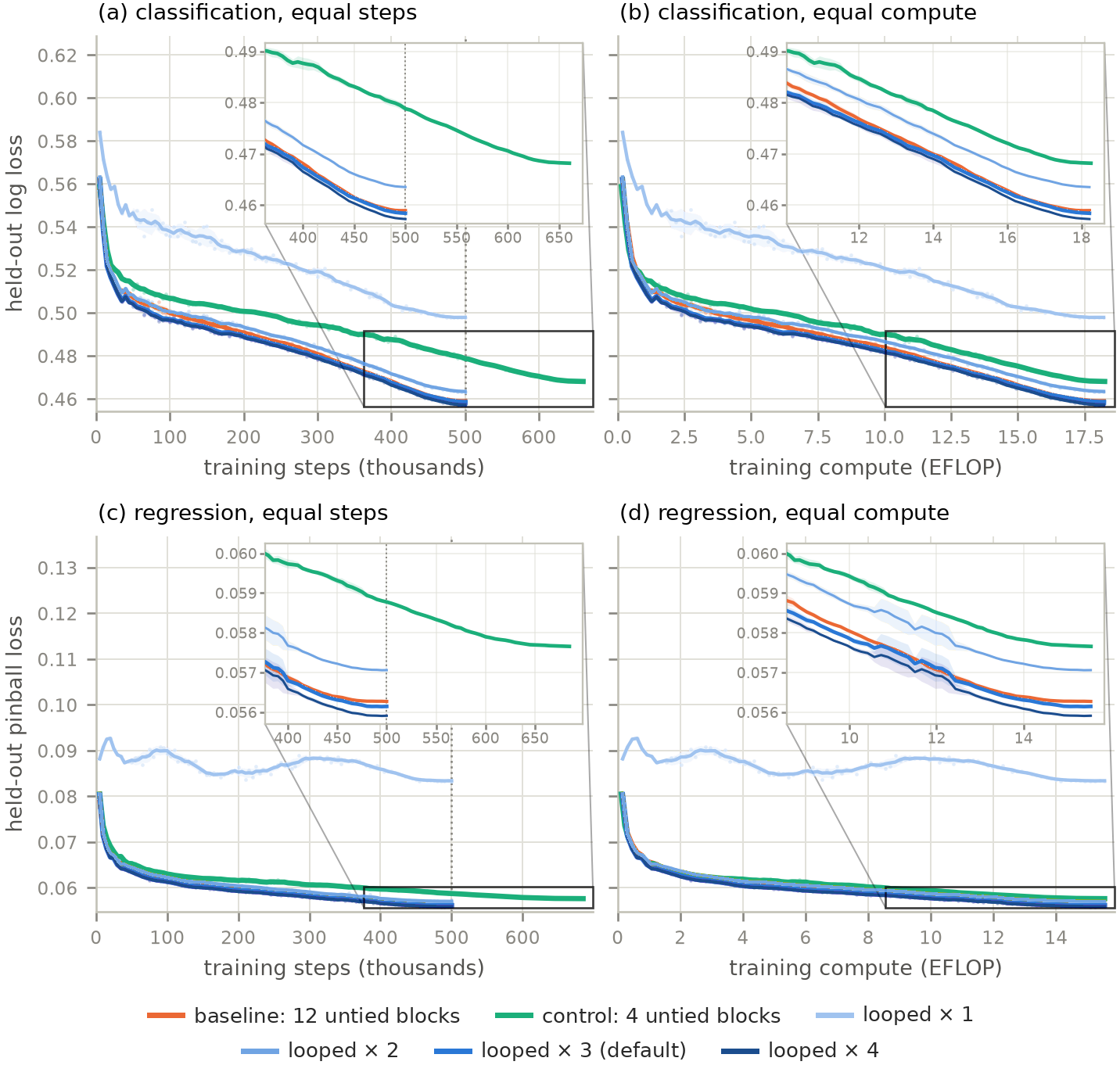}
  \caption{\textbf{Held-out loss during stage~1}, against training steps (a, c) and training compute (b, d), for classification (query log loss) and regression (pinball loss). The control trains for more, cheaper steps to match compute; insets enlarge the end of training. In both tasks, \method{} at $k=3$ tracks the baseline throughout training, and the control ends above every reading with $k\geq2$.}
  \label{fig:heldout}
\end{figure}

\subsection{Classification at Intermediate Budgets}
\label{app:intermediate}

Figure~\ref{fig:equal_compute_intermediate} compares the three stage-1 models at two intermediate budgets of equal training compute. The control trails \method{} at $k=3$ at both budgets. \method{} and the baseline trade places: at $3.6$~EFLOP, \method{} leads on TabArena-Lite ($1215$ vs.\ $1197$) and TALENT ($82.65\%$ vs.\ $82.58\%$); at $7.9$~EFLOP, the baseline leads on both ($1276$ vs.\ $1248$; $83.15\%$ vs.\ $82.90\%$). By the end of training at $18.2$~EFLOP the two are equivalent (Section~\ref{sec:matched}).

\begin{figure}[!htbp]
  \centering
  \includegraphics[width=\linewidth]{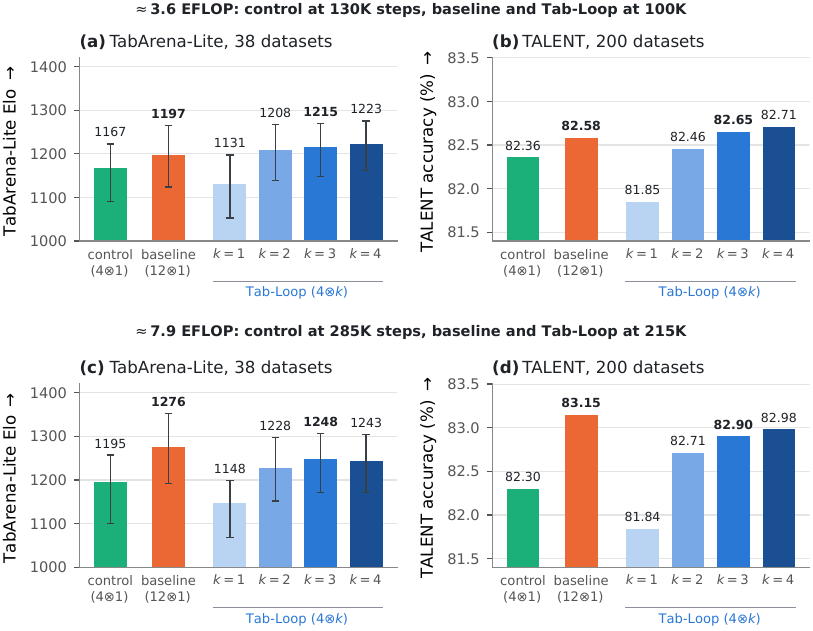}
  \caption{\textbf{Equal-compute comparison at intermediate budgets.} \textbf{Top:} control at 130K steps, baseline and \method{} at 100K steps ($\approx 3.6$~EFLOP). \textbf{Bottom:} control at 285K steps, baseline and \method{} at 215K steps ($\approx 7.9$~EFLOP). (a, c) TabArena-Lite (38 datasets), Elo with 95\% bootstrap intervals. (b, d) TALENT (200 datasets), mean test accuracy. \method{} is read at $k\in\{1,2,3,4\}$; $k=3$ matches the baseline's 12 block applications.}
  \label{fig:equal_compute_intermediate}
\end{figure}

\subsection{Regression After Full Training (Data Matched)}
\label{app:reg_results}
Figure~\ref{fig:bars}c and~d and Table~\ref{tab:main}b report the regressors after all three stages, each continued from its 500K-step stage-1 checkpoint on the same tables. On TALENT regression, \method{} at $k=3$ and $k=4$ wins more often than the control (53\% and 58\% against 41\%) but less often than the baseline (64\%), and at $k=2$ it wins least (33\%). On TabArena regression, \method{} at $k=3$ and $k=4$ (Elo 1704 and 1709) and the baseline (1721) lie within intervals of several hundred Elo, with the control (1654) and $k=2$ (1643) lower.

% \subsection{Differences Between the Regression Runs}
% \label{app:reg_diff}
% The 12-block regression baseline started its 21 attention output-projection biases from PyTorch's default initialization instead of zero; the other regressors start them at zero. We have not measured the effect of this difference. In regression, ``everything else is shared'' therefore holds up to this initialization, and repeated runs should start all models at zero. The regression data streams are not bit-reproducible across machines, so ``the same stream'' means the same prior and seeds.

% Editorial reconciliation: the k=3, k=4, baseline and control Elo values above follow the current main-results table and overview figure; earlier prose reported 1714, 1716, 1728 and 1661. The k=2 value is retained from the existing appendix.

\clearpage
\section{Test-Time Depth and Training Schedules}
\label{app:depthdial}

\emph{Supports Section~\ref{sec:stage2}.} Figures~\ref{fig:schedule} and~\ref{fig:pareto_small} in Section~\ref{sec:stage2} gives the main results on test-time depth. This appendix gives the stage-1 readings behind Figure~\ref{fig:schedule}a, b (Appendix~\ref{app:ttd}), the spread-out schedule with confidence intervals, the peaks of every curve and the gains from stage~2 to stage~3 (Appendix~\ref{app:stage23}), the inference cost of extra loops and a practical summary, the full comparison with published open-weight models (Appendix~\ref{app:pareto}), our replication of TabICLv2 (Appendix~\ref{app:replication}) and concurrent work (Appendix~\ref{app:concurrent}). The loop schedules themselves are specified in Appendix~\ref{app:setup}.

\subsection{Test-Time Depth of the Stage-1 Checkpoint}
\label{app:ttd}
Figure~\ref{fig:schedule}a, b (Section~\ref{sec:depth}) reads the stage-1 classifier, trained on two to four loops, at one to five loops. Performance rises through the trained range and is flat one loop beyond it: full TabArena Elo is 1288, 1370, 1410, 1424 and 1427 at $k=1$ to 5, against 1396 for the untied baseline, and TALENT accuracy is 82.78\%, 83.83\%, 84.10\%, 84.18\% and 84.15\%, against 83.92\%. These values come from a separate evaluation run from Table~\ref{tab:main} and Figure~\ref{fig:bars}, so they differ from them by up to 3 Elo.

\subsection{Stages 2 and 3: Spread-Out Schedule and Longer Tables}
\label{app:stage23}

Figure~\ref{fig:stage2} complements Figure~\ref{fig:schedule}c, d with 95\% intervals for the spread-out (Poisson) model and with TabArena-Lite. On TALENT, accuracy rises from $83.00\%$ at $k=3$ to $84.42\%$ at $k=16$, between the baseline's stage-2 ($84.39\%$) and stage-3 ($84.46\%$) scores. As on TabArena, the Poisson model at $k\le 4$ is weaker than its own stage-1 checkpoint.

Table~\ref{tab:schedule_peaks} lists the peak of every curve in Figure~\ref{fig:schedule}c, d, the values marked there by large markers; Table~\ref{tab:main}b reports these peaks with the loop count at which each is reached.

\begin{table}[!htbp]
\centering
\small
\caption{\textbf{Peak TabArena Elo and TALENT accuracy of each curve in Figure~\ref{fig:schedule}c, d}, maximized over the loop counts read ($k=1$ to 32 for \method{}). Values are from the evaluation run of Figure~\ref{fig:schedule}; for the stage-1 models they can differ by a few points from Figure~\ref{fig:bars}, which comes from a separate evaluation. Stage~1 is shared by the three \method{} schedules. Stages~2 and~3 are data matched. The \method{} run that keeps $\mathcal{U}\{2,3,4\}$ in stages~2 and~3 was read only at $k=3$ and 4 (Table~\ref{tab:main}b).}
\label{tab:schedule_peaks}
\setlength{\tabcolsep}{5pt}
\begin{tabular}{@{}lccc@{}}
\toprule
Model & Stage 1 (500K) & Stage 2 (+40K) & Stage 3 (+10K) \\
\midrule
\swatch{uniformcol}\ \method{}, uniform, $k\in\{2,\dots,8\}$ & 1422 / 84.18 & 1595 / 84.64 & 1604 / 84.67 \\
\swatch{truncpoiscol}\ \method{}, truncated Poisson, $k\in\{2,\dots,8\}$ & (same) & 1592 / 84.62 & 1613 / 84.64 \\
\swatch{spreadcol}\ \method{}, spread-out Poisson, $k\approx6$--$19$ & (same) & 1550 / 84.49 & 1565 / 84.49 \\
\swatch{baselinecol}\ Untied baseline (27.6M) & 1395 / 83.92 & 1515 / 84.39 & 1536 / 84.46 \\
\swatch{controlcol}\ Untied control (10.4M) & 1336 / 83.46$^{*}$ & 1448 / 83.89 & 1470 / 83.99 \\
\bottomrule
\end{tabular}

\vspace{2pt}
{\footnotesize $^{*}$After 660K steps (compute matched, as in Table~\ref{tab:main}a). Stages~2 and~3 continue the control from its 500K-step checkpoint, like every other model (data matched). Entries are Elo / accuracy (\%).}
\end{table}

The gains of stages~2 and~3 survive the longer tables of stage~3. From stage~2 to stage~3, the TabArena peaks rise for every model: from 1595 to 1604 with the uniform schedule, from 1592 to 1613 with the truncated Poisson, from 1550 to 1565 with the spread-out schedule, and from 1515 to 1536 for the untied baseline.

\begin{figure}[!htbp]
  \centering
  \begin{subfigure}{\linewidth}
    \includegraphics[width=\linewidth]{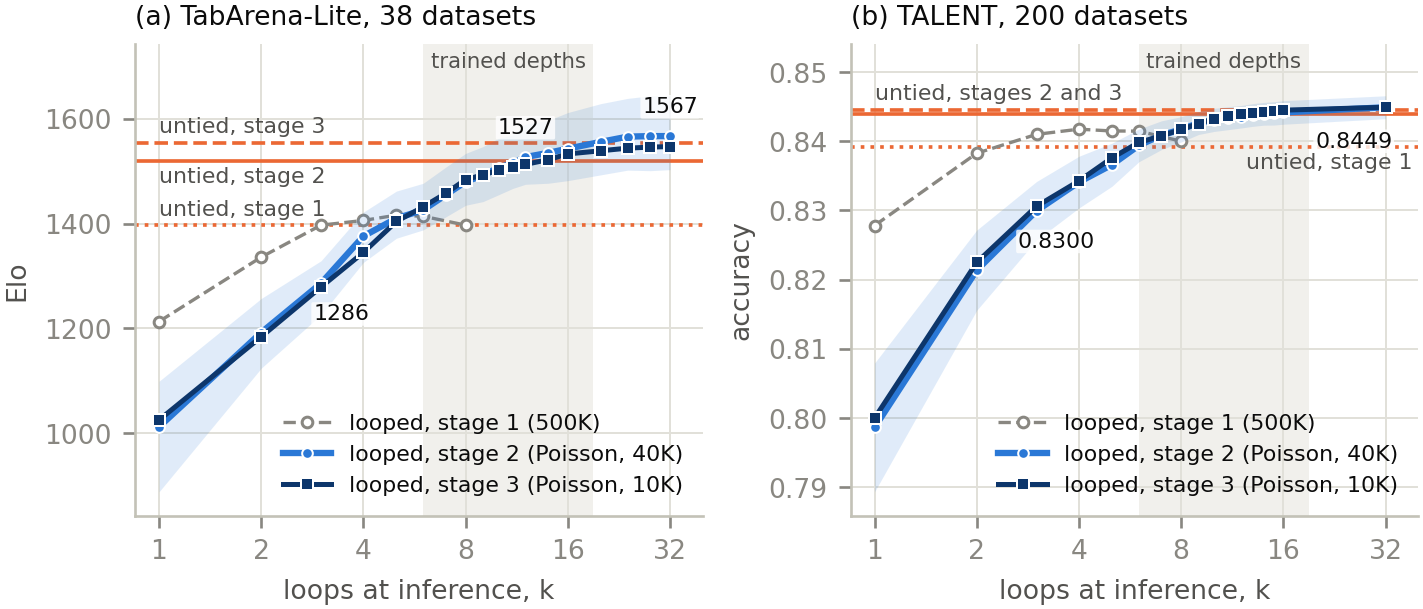}
    \caption{Stage~2 vs.\ stage-1 checkpoint and vs.\ the baseline after stages 1--3.}
  \end{subfigure}\\[6pt]
  \begin{subfigure}{\linewidth}
    \includegraphics[width=\linewidth]{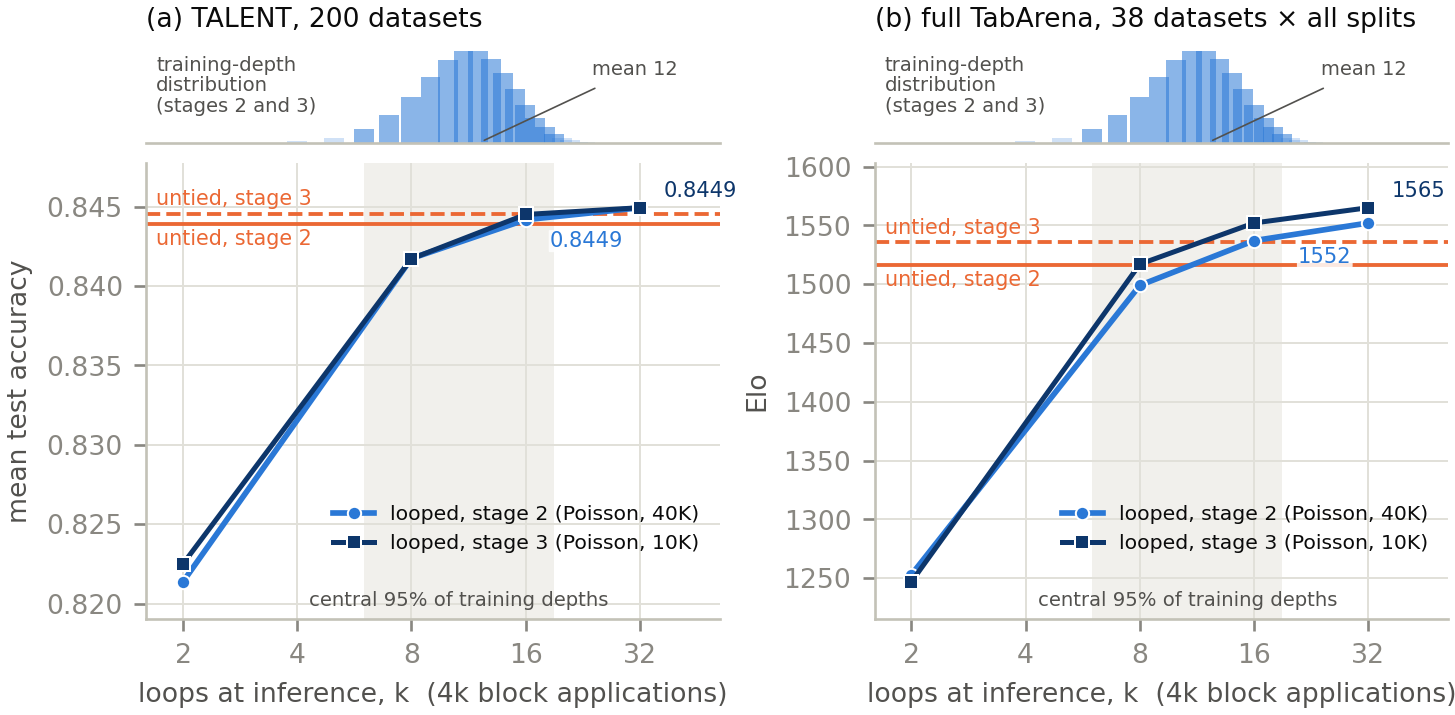}
    \caption{Stage~2 vs.\ stage~3 for both \method{} and the baseline.}
  \end{subfigure}
  \caption{\textbf{Test-time depth after stages~2 and~3 with the spread-out schedule, with 95\% intervals.} Both panels report \method{} trained with $k=\min(2+\mathrm{Poisson}(10),32)$; horizontal lines are the untied baseline at the matching stage. In (a), the left plot is TabArena-Lite and the right TALENT; in (b), the left is TALENT and the right full TabArena. Grey shading marks the central 95\% of training depths; blue bands are 95\% intervals. Axes are logarithmic; $k$ loops cost $4k$ block applications, so $k=16$ and $k=32$ cost $\approx 5.3\times$ and $\approx 10.7\times$ the baseline's ICL block applications. Training compute is not matched (Appendix~\ref{app:train_compute}).}
  \label{fig:stage2}
\end{figure}

\subsection{Comparison With Published Models}
\label{app:pareto}

Figure~\ref{fig:pareto} places our stage-3 models among published open-weight tabular foundation models by parameter count and full TabArena Elo, in classification and regression, over the full Elo range; Figure~\ref{fig:pareto_small} in Section~\ref{sec:frontier} shows the same comparison zoomed to 1000 to 2000 Elo. The published models differ from ours in prior, data and training budget, so only the comparisons among our own models, which share prior and recipe, are controlled.

\begin{figure}[!htbp]
  \centering
  \includegraphics[width=\linewidth]{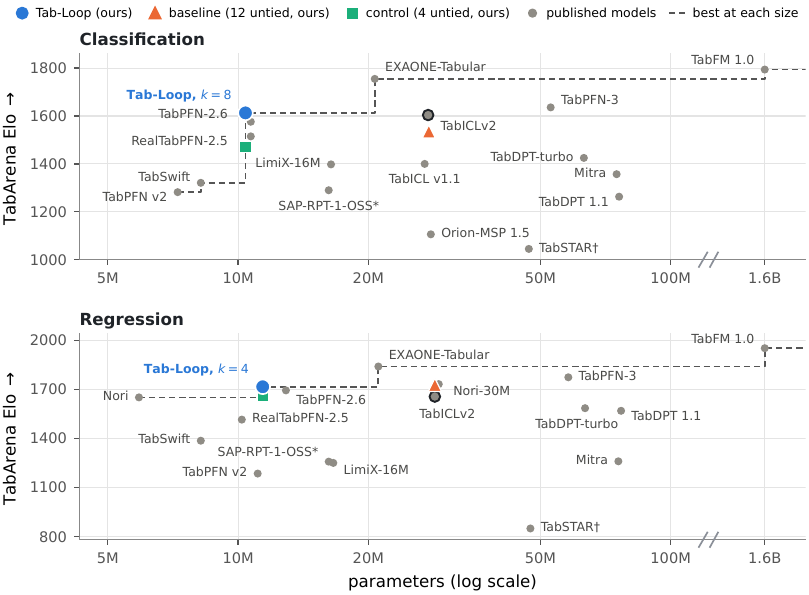}
  \caption{\textbf{Parameters versus Elo among open-weight tabular foundation models, in both tasks.} Full TabArena Elo (default configurations) against parameter count (log scale), for classification (top) and regression (bottom); the ringed point is the official TabICLv2, and \method{} is read at $k=8$ in classification and $k=4$ in regression. Grey: published open-weight models ($^{*}$gated, count from file size; $^{\dagger}$includes its text encoder). Color: our stage-3 models, trained with the same prior and recipe: \method{} (4 weight-shared blocks, truncated-Poisson schedule, read at $k=8$; 10.4M parameters), the untied 12-block replication (27.6M) and the untied 4-block control (10.4M). The dashed line is the Pareto frontier (best Elo at each size); TabFM~1.0 (1.6B parameters) is drawn after the axis break. The horizontal axis counts parameters, not inference compute: at $k=8$, \method{} applies 32 ICL blocks, $2.7\times$ the 12 of the untied baseline. Published models differ from ours in prior, data and training budget, so this comparison is descriptive rather than controlled.}
  \label{fig:pareto}
\end{figure}

\subsection{Our Untied Baseline, the Official TabICLv2 and TabICL}
\label{app:replication}

Our untied baseline is a replication of TabICLv2 trained with our own recipe and compute, not the released checkpoint. Table~\ref{tab:replication} places it next to the official TabICLv2~\citep{qu2026tabiclv2} and its predecessor TabICL~\citep{qu2025tabicl}, using the values of Figure~\ref{fig:pareto}.

\begin{table}[!htbp]
  \centering\small
  \caption{\textbf{Our untied baseline against the released TabICL models} (full TabArena Elo, default configurations). Our models are read after stage~3; \method{} is read at $k=8$ in classification and $k=4$ in regression. $^{\ddagger}$Read off the comparison figure, so approximate.}
  \label{tab:replication}
  \setlength{\tabcolsep}{5pt}
  \begin{tabular}{@{}lcccc@{}}
    \toprule
    Model & ICL blocks & Params & Classification Elo & Regression Elo \\
    \midrule
    TabICL v1.1 (released)              & 12 untied & 27.0M$^{\ddagger}$ & 1400 & -- \\
    TabICLv2 (released)                 & 12 untied & 27.6M & 1604 & 1656$^{\ddagger}$ \\
    Untied baseline (ours, replication) & 12 untied & 27.6M / 28.5M & 1536 & 1728 \\
    \midrule
    \method{} (ours)                    & 4 shared    & 10.4M / 11.4M & 1613 & 1716 \\
    \bottomrule
  \end{tabular}
\end{table}

Three points follow. First, in classification our replication lies between the two released models: 136 Elo above TabICL v1.1 and 68 below the official TabICLv2. We did not try to close this gap, and we cannot attribute it to a single cause; our run follows the published architecture and three-stage curriculum, but the released model may differ from it in the implementation of the prior, in training details and in compute. Second, the gap is not uniform across tasks: in regression our replication scores above the released model. Third, none of this affects the controlled comparisons of the paper, in which \method{}, the baseline and the control share one prior, recipe and compute budget. The released TabICLv2 enters only the descriptive comparison of Section~\ref{sec:frontier}, where \method{} at $k=8$ reaches it in classification with $2.65\times$ fewer parameters and more inference compute.

\subsection{Concurrent Work}
\label{app:concurrent}

Concurrently with our work, LoopICL~\citep{balef2026loopicl} develops a recurrent architecture for tabular ICL. It loops a single shared transformer block with coupled cell and row streams, trains it over varying loop counts with a learned exit gate, and reports results competitive with TabICLv2 on TabArena and TALENT at equal FLOPs with nearly 90\% fewer parameters. The two studies differ in scope. LoopICL proposes one looped design, a single block, and evaluates it on classification. Our study is not limited to one design or task. We train every baseline and control ourselves with the same prior, training recipe and measured training compute, vary the number of distinct blocks from one to twelve and the depth up to 24 block applications, and cover both classification and regression. This lets us attribute differences to weight sharing and to depth, and shows that the finding holds across looped configurations, of which a single looped block, our \shape{1}{12}, is one case.

\clearpage
\apppart{partIII}{Part III\texorpdfstring{\quad}{: }Discussion}
\section{Limitations}
\label{app:limitations}

\emph{Supports Section~\ref{sec:conclusion}.} The limitations below concern the scope of the claims (Appendix~\ref{app:lim_scope}) and the experimental controls (Appendix~\ref{app:lim_controls}). The main-text conclusion also identifies evaluation gaps and the cost of additional loops.

\subsection{Scope of the Claims}
\label{app:lim_scope}

\textbf{One architecture family.} Every model we train is TabICLv2 with a changed ICL stage; the column and row encoders are left untouched and run once. Whether depth rather than distinct parameters also sets the loss of other tabular foundation models, for example those that alternate attention over rows and columns in every layer, is untested.

\textbf{One model size and a bounded range of depths.} The depth--parameter scaling study runs at 3.2~EFLOP per model, about a sixth of the full-scale budget, and the full-scale comparison compares models of a single size (a 27.6M-parameter baseline). The finding that a reused block counts as much as a new one holds up to twice the baseline's depth. At four times the depth, \shape{4}{12} is 0.0006 above 48 untied blocks (Figure~\ref{fig:sweep_analysis}b), within the 0.001 that separates looped and untied models at twelve block applications; we have not tested larger depths. Whether the same picture holds for much larger models and budgets is a hypothesis, not a result of this paper.

% \textbf{The theory is a guide.} The model of Appendix~\ref{app:motivation_full} is linear and its guarantee is asymptotic. It anticipates three of our findings but proves nothing about \method{}.

\subsection{Experimental Controls}
\label{app:lim_controls}

\textbf{Compute is matched only in stage~1.} Stages~2 and~3 are data matched: every model continues from its 500K-step stage-1 checkpoint on the same tables. The loop schedules that train on more loops apply more blocks per step than the baseline, so their training compute in these stages is higher (Appendix~\ref{app:train_compute}); the gains of those schedules over the baseline are therefore not compute matched.

% \textbf{Schedules differ in more than depth.} The spread-out schedule also adds a consistency loss, so its lower and later peak cannot be attributed to its loop distribution alone (Appendix~\ref{app:stage23}).

\textbf{Our baseline is a replication.} The untied baseline follows the published architecture and curriculum but is trained with our recipe and compute; in classification it scores below the released TabICLv2 (Appendix~\ref{app:replication}). The controlled comparisons share one prior and recipe, so this gap does not affect them, but our absolute scores are below what the released recipe reaches. The comparison with published models (Figure~\ref{fig:pareto}) is descriptive, since those models differ from ours in prior, data and training budget.

\end{document}